\documentclass[letterpaper]{article} 
\usepackage{aaai2026}  
\usepackage{times}  
\usepackage{helvet}  
\usepackage{courier}  
\usepackage[hyphens]{url}  
\usepackage{graphicx} 
\usepackage{natbib}  
\usepackage{caption} 
\usepackage{algorithm}
\usepackage{algorithmic}
\usepackage{booktabs}
\usepackage{multirow}
\usepackage{amsmath}
\usepackage{amsfonts}
\usepackage{amssymb}

\graphicspath{{media/}}

\usepackage{float}          
\usepackage[table]{xcolor}  
\usepackage{newtxtt}        
\usepackage{adjustbox}     
\usepackage{colortbl}      
\usepackage{makecell}       
\usepackage{tabularx}      
\usepackage{nicefrac}

\usepackage[most]{tcolorbox}
\tcbset{
  mybox/.style={
    breakable,
    colback=white,
    colframe=black,
    boxrule=0.5pt,
    arc=0mm,
    boxsep=1mm,
    left=1mm,
    right=1mm,
    top=1mm,
    bottom=1mm
  }
}

\lstdefinestyle{mypython}{
    language=Python,
    basicstyle=\ttfamily\color{black},
    commentstyle=\color{black},
    showstringspaces=false,
    breaklines=true,
    frame=none,
    numbers=none,
    aboveskip=0.5em,
    belowskip=0.5em
}

\usepackage{newfloat}
\usepackage{listings}
\DeclareCaptionStyle{ruled}{labelfont=normalfont,labelsep=colon,strut=off} 
\floatstyle{ruled}
\newfloat{listing}{tb}{lst}{}
\floatname{listing}{Listing}
\title{Intention Chain-of-Thought Prompting with Dynamic Routing for \\Code Generation}
\author {
    Shen Li\textsuperscript{\rm 1},
    Li Huang\textsuperscript{\rm 1},
    Shaoxiong Zhan\textsuperscript{\rm 2},
    Weifeng Sun\textsuperscript{\rm 1},
    Tao Yin\textsuperscript{\rm 1},
    Zhongxin Liu\textsuperscript{\rm 3},
    Meng Yan\textsuperscript{\rm 1}\thanks{Corresponding author.}
}
\affiliations {
    \textsuperscript{\rm 1}Chongqing University\\
    \textsuperscript{\rm 2}Tsinghua University\\
    \textsuperscript{\rm 3}Zhejiang University\\
    \{shen.l, yintao\}@stu.cqu.edu.cn, zhansx24@mails.tsinghua.edu.cn\\ liu\_zx@zju.edu.cn, \{lee.h, weifeng.sun, mengy\}@cqu.edu.cn
}

\usepackage{bibentry}

\begin{document}

\maketitle

\begin{abstract}
Large language models (LLMs) exhibit strong generative capabilities and have shown great potential in code generation. Existing chain-of-thought (CoT) prompting methods enhance model reasoning by eliciting intermediate steps, but suffer from two major limitations: First, their uniform application tends to induce overthinking on simple tasks. Second, they lack intention abstraction in code generation, such as explicitly modeling core algorithmic design and efficiency, leading models to focus on surface-level structures while neglecting the global problem objective. Inspired by the cognitive economy principle of engaging structured reasoning only when necessary to conserve cognitive resources, we propose RoutingGen, a novel difficulty-aware routing framework that dynamically adapts prompting strategies for code generation. For simple tasks, it adopts few-shot prompting; for more complex ones, it invokes a structured reasoning strategy, termed Intention Chain-of-Thought (ICoT), which we introduce to guide the model in capturing task intention, such as the core algorithmic logic and its time complexity. Experiments across three models and six standard code generation benchmarks show that RoutingGen achieves state-of-the-art performance in most settings, while reducing total token usage by 46.37\% on average across settings. Furthermore, ICoT outperforms six existing prompting baselines on challenging benchmarks. 
\end{abstract}

\begin{links}
    \link{Code}{https://github.com/Guai001/RoutingGen}
\end{links}

\begin{figure*}[htbp]
\centering
\includegraphics[width=2\columnwidth]{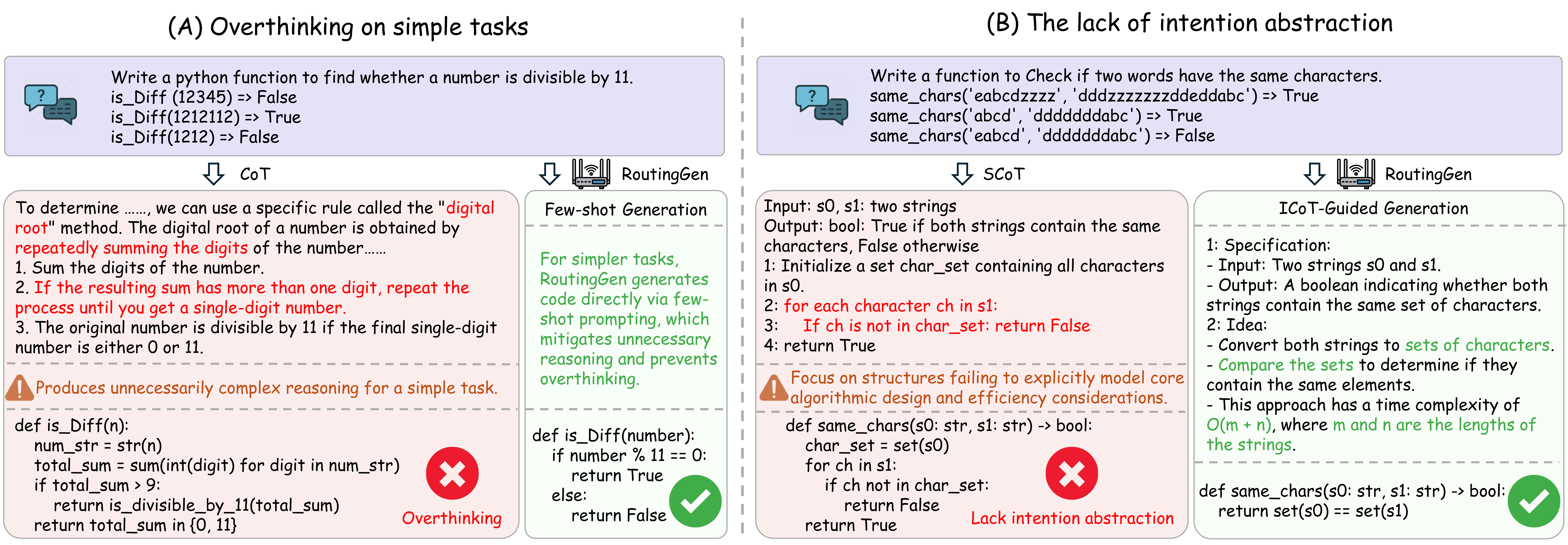} 
\caption{Limitations of existing CoT prompting methods in code generation. (A) Overthinking on simple tasks due to uniform application of structured prompting at the functional code level. (B) The lack of intention abstraction in code generation, such as algorithmic design and efficiency modeling.}
\label{fig0}
\end{figure*}

\section{Introduction}
Code generation focuses on translating user requirements into executable programs and is regarded as a core task in software engineering~\cite{chen2021evaluatinglargelanguagemodels,austin2021programsynthesislargelanguage}. LLMs exhibit strong generative capabilities and have shown great potential in this task. This performance stems from their ability to simulate complex reasoning processes in structured problem solving~\cite{wei2022chain,zheng2025makeslargelanguagemodels,tian2025codehalu}. However, a critical gap remains between this reasoning capability and its application in code generation: LLMs tend to produce syntactically correct programs that fail to align with the task intention, resulting in functionally incorrect outputs~\cite{cobbe2021trainingverifierssolvemath, jiang2024self, chen2023program}.

To address these challenges, recent studies have explored prompting strategies that guide models to generate intermediate steps. Scratchpads~\cite{nye2021workscratchpadsintermediatecomputation} introduced the idea of showing intermediate computations. CoT prompting~\cite{wei2022chain} then generalized this idea across tasks, followed by self-consistency decoding~\cite{wang2023selfconsistencyimproveschainthought}, which enhances consistency through multiple sampled traces. To better align model reasoning with programming tasks, subsequent work has proposed code-centric prompting strategies such as Program-of-Thought ~\cite{chen2023program}, CodeCoT~\cite{huang2024codecottacklingcodesyntax}, and self-planning~\citep{jiang2024self}. Despite their progress, these methods reveal two key limitations. First, their uniform application across functional-level programming problems often induces overthinking on simple tasks, resulting in disorganized logic and reduced accuracy (Figure~\ref{fig0}(A)). Second, they lack intention abstraction in code generation, failing to explicitly model core algorithmic design and efficiency considerations. As a result, models tend to focus on structural correctness while neglecting the intended task objective (Figure~\ref{fig0}(B)).

Dual-process theories of human cognition describe two complementary systems: System 1 supports rapid and intuitive responses to simple or familiar problems, while System 2 is engaged for deliberate and structured reasoning when tasks are complex~\citep{kahneman2011thinking}. This adaptive mechanism reflects the principle of cognitive economy,  which promotes conserving cognitive resources by activating structured deliberation only when necessary~\citep{stanovich2000individual}.   Inspired by this principle, we propose RoutingGen, a difficulty-aware dynamic routing framework that employs a classifier to estimate problem difficulty and dynamically selects appropriate prompting strategies. For simpler tasks,  RoutingGen generates code directly via few-shot prompting, which mitigates unnecessary reasoning and prevents overthinking. For more complex problems, it invokes a structured reasoning strategy, termed \textit{Intention Chain-of-Thought} (ICoT), which we introduce to guide the model in capturing task intent. Specifically, ICoT comprises two components: a \textit{Specification} element that defines the input-output constraints, and an \textit{Idea} element that captures the core algorithmic logic and estimates time complexity. This decomposition reflects the classic problem-solving strategy of separating task comprehension from solution design~\cite{polya1945solve}, whose importance has been further underscored by recent advances in mathematical reasoning with LLMs~\cite{wang2023planandsolvepromptingimprovingzeroshot}. As a result, this intention representation helps steer code generation toward solutions that preserve structural guidance while explicitly modeling the task’s functional requirements. 

We evaluate RoutingGen and ICoT across three models and six standard code generation benchmarks. RoutingGen achieves state-of-the-art performance in most settings while reducing total token usage by 46.37\% on average. Additionally, ICoT consistently outperforms six prompting baselines on challenging benchmarks. Furthermore, ablation results show that RoutingGen demonstrates robustness to variations in the difficulty classification model and that both the Specification and Idea stages contribute to ICoT’s effectiveness.

Our contributions are threefold:
\begin{itemize} 
    \item We focus on the issue of overthinking on simple tasks due to uniform application of structured prompting at the functional code level, and identify a core limitation in existing methods: the lack of intention abstraction in code generation, such as algorithmic design and efficiency modeling.
    
    \item We propose \textit{RoutingGen}, a novel difficulty-aware routing framework that dynamically adapts prompting strategies for code generation. For simple tasks, it adopts few-shot prompting; for more complex ones, it invokes a structured reasoning strategy, termed \textit{Intention Chain-of-Thought} (ICoT), which we introduce to guide the model in capturing task intent, including core algorithmic logic and estimated time complexity.

    \item We empirically validate that RoutingGen achieves state-of-the-art performance in most settings while substantially reducing token usage. Additionally, ICoT consistently outperforms six prompting baselines on challenging benchmarks.
\end{itemize}

\begin{figure*}[htbp]
\centering
\includegraphics[width=1.9\columnwidth]{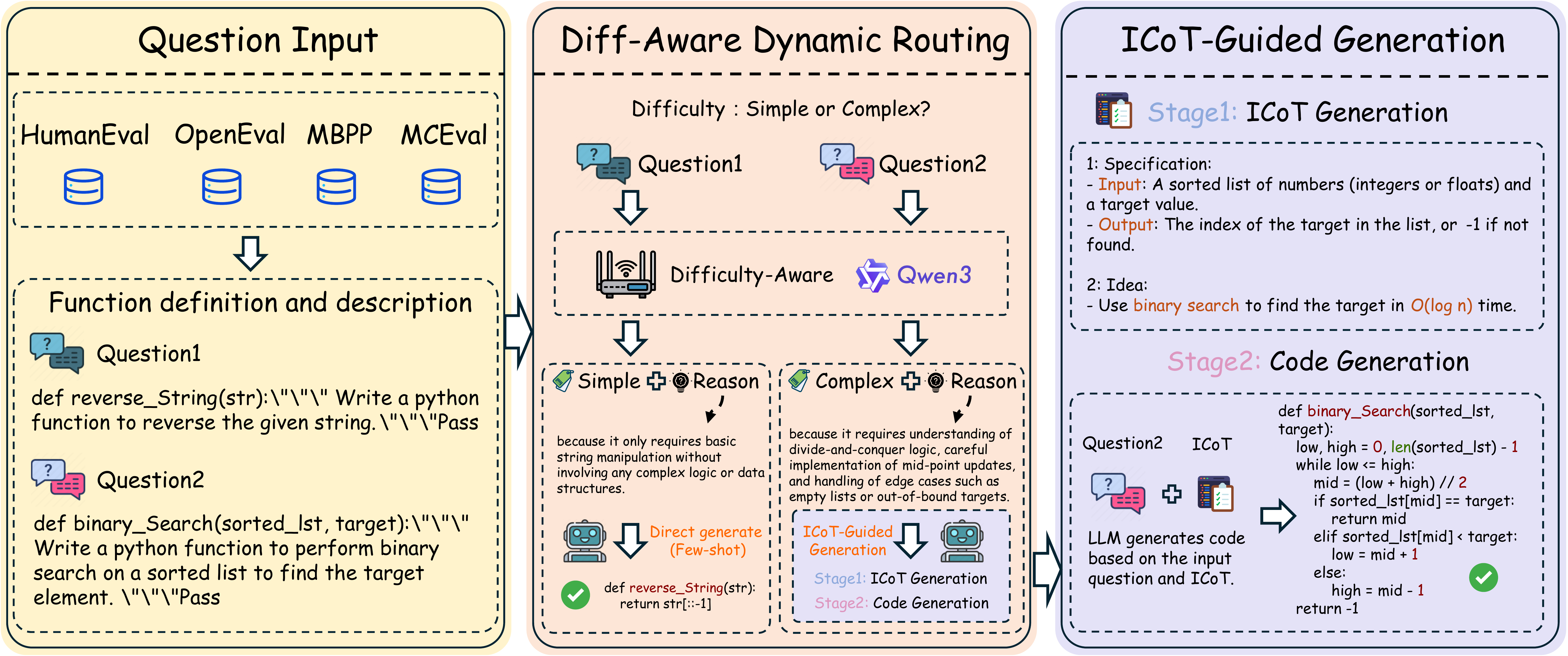} 
\caption{The RoutingGen Framework for Difficulty-Aware Dynamic Routing. For simple problems, it adopts few-shot prompting, while for more challenging cases, it leverages a structured reasoning strategy we propose, termed ICoT, which captures task intention, including the core algorithmic logic and time complexity.}
\label{fig2}
\end{figure*}

\section{Related Work}

The impressive reasoning capabilities exhibited by LLMs have led to extensive research on prompting strategies~\cite{wei2022chain, kojima2022large, kaplan2020scaling}. A seminal contribution in this direction is the CoT prompting method~\cite{wei2022chain, kojima2022large}, which guides models to articulate intermediate reasoning steps. This approach has inspired a range of prompting methods designed to make reasoning more explicit and structured. Representative strategies include enhancing robustness through multi-path sampling as in Self-Consistency~\cite{wang2023selfconsistencyimproveschainthought}, applying structured planning in Tree-of-Thoughts~\cite{yao2023tree}, incorporating step-wise decomposition~\cite{zhou2023leasttomostpromptingenablescomplex}, and enabling self-correction via reflexion~\cite{shinn2023reflexion}. However, the uniform application of these methods tends to induce overthinking on simple tasks.

To better align model reasoning with programming tasks, recent work has proposed code-centric prompting strategies, such as leveraging abstract syntax trees~\cite{yin2017syntacticneuralmodelgeneralpurpose}, incorporating self-planning~\cite{jiang2024self}, employing structured reasoning frameworks~\cite{li2025structured}, integrating execution-time validation with compiler feedback~\cite{gao2023pal, zelikman2024self}, and leveraging in-context learning to organize requirements observed in descriptions and to extrapolate unexpressed requirements~\cite{han-etal-2024-archcode}. In contrast to their approach, we place greater emphasis on the abstraction of task intention in code generation, which helps steer code generation toward solutions that preserve structural guidance while explicitly modeling the task’s functional requirements. 

Current studies on routing techniques mainly focus on demonstration selection or cost efficiency. For example, Auto-CoT automates prompt construction through clustering~\cite{zhang2022automaticchainthoughtprompting}, while other studies route queries to different models or adapt sampling strategies~\cite{varangotreille2025doinglesssurveyrouting, wang-etal-2025-make}. Our framework differs in that it is motivated by the principle of cognitive economy and uniquely routes between distinct reasoning strategies based on task complexity.

\section{Methodology}

\label{sec:RoutingGen}

In this section, we introduce \textbf{RoutingGen}, a novel difficulty-aware routing framework that dynamically adapts prompting strategies based on problem difficulty in code generation. As illustrated in Figure~\ref{fig2}, the overall workflow consists of two key components following the initial input: a Difficulty-Aware Routing module that dynamically assigns each problem to a suitable generation strategy, and an ICoT-Guided Generation process for complex tasks, which guides the model in capturing task intent.

\subsection{Difficulty-Aware Dynamic Routing}
RoutingGen leverages $\mathcal{M}_{\text{cls}}$ (Qwen3-8B) as a difficulty-aware classifier to steer the selection of prompting strategies for a given input problem $q$. Conditioned on a carefully designed prompt, the classifier assigns $q$ to one of two difficulty levels from the label space $\mathcal{L} = \{\text{Simple}, \text{Complex}\}$ and generates a textual rationale $r$ explaining its decision. Formally, this classification step is defined as:
\begin{equation}
(d^*, r^*) = \operatorname*{argmax}_{(d, r) \text{ s.t. } d \in \mathcal{L}} P_{\mathcal{M}_{\text{cls}}}(d, r \mid q, T_{\text{cls}})
\label{eq:difficulty_classifier}
\end{equation}
where $T_{\text{cls}}$ is the classification prompt. Here, \( d^* \) denotes the assigned difficulty label, and \( r^* \) is the corresponding rationale produced by the classifier.

Based on the assigned difficulty label $d^*$, RoutingGen subsequently applies the corresponding generation strategy tailored to the task complexity:
\begin{equation}
\mathcal{G}_{\text{strategy}} = f(d^*) =
\begin{cases}
\mathcal{G}_{\text{Direct}} & \text{if } d^* = \mathit{Simple} \\
\mathcal{G}_{\text{ICoT}} & \text{if } d^* = \mathit{Complex}\
\end{cases}
\label{eq:routing_policy}
\end{equation}
we use a model $\mathcal{M}_{\text{gen}}$ to perform both intention and code generation throughout the framework. For problems classified as \textit{Simple}, RoutingGen applies a direct, low-cost generation strategy $\mathcal{G}_{\text{direct}}$ based on few-shot prompting. The model generates a set of $n$ candidate code solutions by sampling from its conditional distribution:
\begin{equation}
C_i^* \sim P_{\mathcal{M}_{\text{gen}}}(\cdot \mid q, T_{\text{Direct}}) \quad \text{for } i = 1, \dots, n
\label{eq:direct_generation}
\end{equation}
where $T_{\text{direct}}$ is a predefined few-shot prompt template. The resulting set of outputs is denoted as: 
\begin{equation}
\mathcal{G}_{\text{Direct}}(q) = \{C_1^*, \dots, C_n^*\}
\label{eq:Direct_generate}
\end{equation}
this constitutes the output of the direct generation strategy for simple problems.

In contrast, for more challenging cases classified as \textit{Complex}, RoutingGen employs a structured reasoning strategy, which we term \textbf{ICoT-Guided Generation}. We detail this two-stage process in the following subsection.

\subsection{ICoT-Guided Generation}

The ICoT-guided generation process, illustrated in the right panel of Figure~\ref{fig2}, comprises two stages. The first stage, \textit{ICoT Generation}, explicitly models the task intention using a specification-and-idea structure that captures both the functional requirements and the global algorithmic strategy, including core logic and efficiency considerations. The second stage, \textit{Code Generation}, directs the model to generate code conditioned on both the input question and the generated ICoT.

\paragraph{Stage 1: ICoT Generation}
In this stage, the model is prompted with the input problem to generate a diverse set of $n$ candidate \textit{ICoT} instances. Each instance is a structured pair of a \textit{Specification} and an \textit{Idea}. This generation process leverages stochastic decoding techniques (e.g., nucleus sampling) applied over the model's conditional distribution. The resulting set is denoted as $\mathcal{R}_{\text{ICoT}} = \{ \text{ICoT}_i \}_{i=1}^n$, where each instance is sampled as:
\begin{equation}
\text{ICoT}_i \sim P_{\mathcal{M}_{\text{gen}}}(\cdot \mid q, T_{\text{ICoT}}^{(1)}) \quad \text{for } i=1, \dots, n
\label{eq:ICoT_stage1_sampling}
\end{equation}
here, $T_{\text{ICoT}}^{(1)}$ denotes the prompt used for \textit{Intention generation}, and $P_{\mathcal{M}_{\text{gen}}}$ is the conditional distribution induced by $\mathcal{M}_{\text{gen}}$.
Each $\text{ICoT}_i$ is a unified structured output comprising a \textit{Specification} $S_i$ and an \textit{Idea} $I_i$, i.e., $\text{ICoT}_i = (S_i, I_i)$. These pairs are generated jointly in a single decoding pass.

\paragraph{Stage 2: Code Generation}
In the second stage, each $\text{ICoT}_i = (S_i, I_i)$ from $\mathcal{R}_{\text{ICoT}}$ is used to generate a corresponding code solution $C_i^*$ via greedy decoding. The model is conditioned on both the input problem $q$ and its associated $\text{ICoT}_i$, guiding \textit{code generation} aligned with the task’s functional objective:

\begin{equation}
C_i^* = \text{GreedyDecode}_{\mathcal{M}_{\text{gen}}}(q, \text{ICoT}_i, T_{\text{ICoT}}^{(2)}) \quad \text{for } i = 1, \dots, n
\label{eq:ICoT_stage2_greedy}
\end{equation}
where $T_{\text{ICoT}}^{(2)}$ denotes the prompt template for code generation. Each implementation $C_i^*$ is the unique token sequence deterministically generated by $\mathcal{M}_{\text{gen}}$ under greedy decoding.

The final output is a set of $n$ candidate code completions derived from the ICoT-guided process:
\begin{equation}
\mathcal{G}_{\text{ICoT}}(q) = \{C_1^*, \dots, C_n^*\}
\label{eq:ICoT_overall}
\end{equation}

\section{Experiment Setup}
\subsection{Benchmarks}
Following recent work in LLM evaluation~\cite{openai2024gpt4technicalreport,jiang2024self,yang2025qwen3technicalreport}, we evaluate on six widely used code generation benchmarks. \textbf{HumanEval}\cite{chen2021evaluatinglargelanguagemodels} contains 164 Python problems with reference implementations and test cases. \textbf{MBPP-sanitized}\cite{austin2021programsynthesislargelanguage} includes 427 verified tasks with three tests per instance. \textbf{HumanEval-ET} and \textbf{MBPP-ET}\cite{dong2025codescore} extend the original sets with around 100 edge-case tests per problem. \textbf{OpenEval}\cite{yang2024chain} comprises 178 challenging problems from AVATAR, with manually written test cases. \textbf{McEval}~\cite{chai2024mceval} is a multilingual benchmark; we use its Python subset of 50 problems, adopting the difficulty labels from the original release.

\subsection{Large Language Models}
\paragraph{Difficulty-Aware Classifier.}
We employ Qwen3-8B as the difficulty classifier, selected for its strong performance on code reasoning tasks and competitive alignment with human preferences~\cite{yang2025qwen3technicalreport}.

\paragraph{Code Generation.} We evaluate our method on three high-performing models specialized for code generation. \textbf{Qwen2.5-Coder-3B-Instruct}~\cite{hui2024qwen2} is a 3B-parameter instruction-tuned model in the Qwen series (formerly CodeQwen), demonstrating strong performance on code generation, mathematical reasoning, and general problem solving. \textbf{DeepSeek-Coder-6.7B-Instruct}~\cite{guo2024deepseek} is a state-of-the-art open-source model that demonstrates robust results across multiple programming languages and standard benchmarks. We also include \textbf{DeepSeek-V3}~\cite{liu2024deepseek}, a Mixture-of-Experts model with 671B total parameters, which achieves performance competitive with or surpassing proprietary LLMs.

\subsection{Baselines}
\textbf{Self-CoT} \cite{yang2024chain} encourages the model to generate natural language reasoning before producing the final output. \textbf{Zero-shot-CoT} \cite{kojima2022large}, denoted as ZS-CoT in our results, is a zero-shot prompting approach that guides multi-step reasoning through a simple prefix. \textbf{Self-planning} \cite{jiang2024self}, denoted as SP in our results, adopts a two-stage framework, where the model first generates a numbered subtask plan and then uses it to guide code generation. \textbf{SCoT} \cite{li2025structured} incorporates sequential, branching, and looping structures into natural language reasoning to align prompts with program logic.

\begin{table*}[tb]
\centering
\small
\setlength{\tabcolsep}{1mm}
\begin{tabular}{@{}ccccccc@{}}
\toprule
 \textbf{Method} & \textbf{HumanEval} & \textbf{HumanEval-ET} & \textbf{MBPP-sanitized} & \textbf{MBPP-ET} & \textbf{OpenEval} & \textbf{McEval} \\
\midrule

\multicolumn{7}{l}{\textbf{Qwen2.5-Coder-3B-Instruct}} \\ 
\midrule
zero-shot 
& 75.49\% & 67.29\% & 61.42\%
& 44.12\% & 35.06\% & 26.80\% \\

few-shot 
& 72.80\% (-3.56\%) & 65.91\% (-2.05\%) 
& \underline{68.74\% (+11.92\%)} & \underline{48.20\% (+9.25\%)} 
& 34.75\% (-0.88\%) & 31.10\% (+16.04\%) \\

Self-CoT 
& 71.55\% (-5.22\%) & 64.18\% (-4.62\%) 
& 66.01\% (+7.47\%) & 46.51\% (+5.42\%) 
& 34.13\% (-2.65\%) & 23.50\% (-12.31\%) \\

ZS-CoT
& \underline{75.70\% (+0.28\%)} & \underline{67.99\% (+1.04\%)} 
& 66.73\% (+8.65\%) & 47.24\% (+7.07\%) 
& 35.42\% (+1.03\%) & 26.20\% (-2.24\%) \\

SP
& 72.84\% (-3.51\%) & 64.02\% (-4.86\%) 
& 53.69\% (-12.59\%) & 36.93\% (-16.30\%) 
& \underline{35.65\% (+1.68\%)} & 25.30\% (-5.60\%) \\

SCoT
& 65.27\% (-13.54\%) & 58.60\% (-12.91\%) 
& 61.62\% (+0.33\%) & 41.87\% (-5.10\%) 
& 33.57\% (-4.25\%) & \underline{35.10\% (+30.97\%)} \\
\cmidrule{2-7} 

RoutingGen
& \textbf{76.65\%} \textbf{(+1.54\%)} & \textbf{69.02\%} \textbf{(+2.57\%)} 
& \textbf{68.84\%} \textbf{(+13.71\%)} & \textbf{49.33\%} \textbf{(+11.81\%)}
& \textbf{35.76\%} \textbf{(+2.00\%)} & \textbf{35.30\%} \textbf{(+31.72\%)} \\ 

ICoT 
& \textbf{77.10\%} \textbf{(+2.13\%)} & \textbf{69.73\%} 
\textbf{(+3.63\%)} & \textbf{69.11\%} \textbf{(+12.52\%)} 
& \textbf{48.58\%} \textbf{(+10.11\%)} & \textbf{35.70\%} 
\textbf{(+1.83\%)} & \textbf{38.90\%} \textbf{(+45.15\%)} \\ 

\midrule
\multicolumn{7}{l}{\textbf{DeepSeek-Coder-6.7B-Instruct}} \\ 
\midrule
zero-shot 
& 45.95\% & 39.60\% & 46.08\% 
& 31.10\% & 16.74\% & 23.80\% \\

few-shot 
& \underline{72.68\% (+58.17\%)} & \underline{64.27\% (+62.30\%)}
& \textbf{73.40\% (+59.29\%)} & \underline{51.64\% (+66.05\%)} 
& \underline{37.89\% (+126.34\%)} & 39.30\% (+65.13\%) \\

Self-CoT
& 68.29\% (+48.62\%) & 60.27\% (+52.20\%)
& 36.70\% (-20.36\%) & 23.95\% (-22.99\%) 
& 35.34\% (+111.11\%) & 38.70\% (+62.61\%) \\

ZS-CoT
& 63.66\% (+38.54\%) & 55.58\% (+40.35\%) 
& 38.01\% (-17.51\%) & 24.75\% (-20.42\%) 
& 34.35\% (+105.20\%) & 34.80\% (+46.22\%) \\

SP 
& 58.90\% (+28.18\%) & 52.47\% (+32.50\%) 
& 49.19\% (+6.75\%) & 33.72\% (+8.42\%)
& 25.17\% (+50.36\%) & 30.50\% (+28.15\%) \\

SCoT 
& 70.79\% (+54.06\%) & 62.87\% (+58.76\%) 
& 67.70\% (+46.92\%) & 46.58\% (+49.77\%) 
& 36.91\% (+120.49\%) & \underline{40.90\% (+71.85\%)} \\
\cmidrule{2-7} 

RoutingGen 
& \textbf{73.51\%} \textbf{(+59.97\%)} & \textbf{64.76\%} \textbf{(+63.53\%)} 
& \underline{72.30\% (+56.90\%)} & \textbf{51.73\%} \textbf{(+66.33\%)}
& \textbf{38.74\%} \textbf{(+131.42\%)} & \textbf{41.00\%} \textbf{(+72.27\%)} \\ 

ICoT 
& \textbf{73.14\%} \textbf{(+59.17\%)} & \textbf{65.03\%} \textbf{(+64.22\%)}
& \underline{71.24\% (+54.60\%)} & \underline{50.85\% (+63.50\%)} 
& \textbf{38.23\%} \textbf{(+128.38\%)} & \textbf{42.90\%} \textbf{(+80.25\%)} \\ 

\midrule
\multicolumn{7}{l}{\textbf{DeepSeek-V3-671B}} \\ 
\midrule

zero-shot 
& 85.61\% & 77.93\% & {88.99\%} 
& 63.04\% & 46.97\% & 33.20\% \\

few-shot 
& 84.76\% (-0.99\%) & 75.73\% (-2.82\%) 
& \underline{89.23\% (+0.27\%)} & \textbf{63.23\% (+0.30\%)} 
& 42.81\% (-8.86\%) & 51.60\% (+55.42\%) \\

Self-CoT 
& \underline{91.34\% (+6.69\%)} & 81.71\% (+4.85\%) 
& 83.98\% (-5.63\%) & 60.09\% (-4.68\%) 
& 45.51\% (-3.11\%) & \underline{64.40\% (+93.98\%)} \\

ZS-CoT
& 90.85\% (+6.12\%) & \underline{82.32\% (+5.63\%)}
& 84.92\% (-4.57\%) & 60.42\% (-4.16\%) 
& 41.01\% (-12.69\%) & 54.00\% (+62.65\%) \\

SP 
& 80.37\% (-6.12\%) & 73.17\% (-6.11\%) 
& 80.52\% (-9.52\%) & 55.93\% (-11.28\%) 
& 42.81\% (-8.86\%) & 51.60\% (+55.42\%) \\

SCoT 
& 90.98\% (+6.27\%) & 81.34\% (+4.38\%) 
& 79.30\% (-10.89\%) & 54.71\% (-13.21\%)
& \underline{47.08\% (+0.23\%)} & 60.00\% (+80.72\%) \\
\cmidrule{2-7} 

RoutingGen 
& \textbf{91.83\%} \textbf{(+7.27\%)} & \underline{82.07\% (+5.31\%)} 
& \textbf{90.21\%} \textbf{(+1.37\%)} & \underline{63.14\% (+0.16\%)} 
& \textbf{47.98\%} \textbf{(+2.15\%)} & \textbf{65.20\%} \textbf{(+96.39\%)} \\

ICoT 
& \textbf{92.07\%} \textbf{(+7.55\%)} & \textbf{82.68\%} \textbf{(+6.10\%)} 
& 80.70\% (-9.31\%) & 56.25\% (-10.77\%) & \textbf{47.30\%}\textbf{(+0.70\%)}
& \textbf{67.20\%} \textbf{(+102.41\%)} \\

\bottomrule
\end{tabular}
\caption{Pass@1 comparison across three models and six code generation benchmarks. Results cover two direct generation baselines, four structured prompting baselines, and the proposed RoutingGen and ICoT. Bold and underline indicate the best and second-best among our proposed methods and all baselines. Parentheses indicate relative improvements over zero-shot.}
\label{tab:result_pass@}
\end{table*}

\subsection{Evaluation Metrics}

We evaluate our method along two primary axes: generative effectiveness and computational efficiency.

\paragraph{Effectiveness.}
We measure effectiveness using the Pass@k metric~\cite{Li_2022}, which estimates the probability that at least one correct program is included when selecting $k$ candidates (where $k \leq n$) uniformly at random without replacement from a set of $n$ generated candidates, among which $c$ candidates pass all test cases. The unbiased estimator of Pass@k is defined as:
\begin{equation}
\text{Pass@}k := \mathrm{E}_{\text{problem}} \left[1 - \frac{\binom{n - c}{k}}{\binom{n}{k}} \right]
\end{equation}

\paragraph{Efficiency.}
To quantify the computational efficiency promised by our framework, we measure the total number of tokens processed per problem. The total cost for a problem $q$ is determined by the routing decision $d^*$ and defined as the sum of its input and output token counts.
For problems routed to the direct generator ($\mathcal{G}_{\text{direct}}$), the cost comprises the tokens in the single input prompt and the cumulative tokens of all $n$ generated code candidates. For problems routed to ICoT ($\mathcal{G}_{\text{ICoT}}$), the cost is aggregated across its two-stage process. The first stage cost includes one input prompt and the $n$ resulting intention outputs. The second stage cost includes $n$ separate input prompts (one for each intention) and their corresponding single code outputs.
Formally, let $C(\cdot)$ denote the token count of a text sequence. The input and output costs are defined as:

\begin{equation}
\label{eq:token_cost_in}
\mathcal{C}_{\text{in}}(q) = 
\begin{cases}
C(T_{\text{direct}}(q)), & \mathit{Simple} \\
C(T_{\text{ICoT}}^{(1)}(q)) 
+ \sum\limits_{i=1}^{n} C(T_{\text{ICoT}}^{(2)}(q, \text{ICoT}_i)), & \mathit{Complex}
\end{cases}
\end{equation}

\begin{equation}
\label{eq:token_cost_out}
\mathcal{C}_{\text{out}}(q) = 
\begin{cases}
\sum\limits_{i=1}^{n} C(\text{code}_i), & \mathit{Simple} \\
\sum\limits_{i=1}^{n} C(\text{ICoT}_i) 
+ \sum\limits_{i=1}^{n} C(\text{code}_i), & \mathit{Complex}
\end{cases}
\end{equation}
where $T_{\text{direct}}$ and $T_{\text{ICoT}}^{(1,2)}$ are the respective prompt templates, $\text{ICoT}_i$ is the $i$-th generated ICoT, and $\text{code}_i$ is the final code snippet generated from the corresponding path.

The total token usage per problem is:
\begin{equation}
\label{eq:token_total}
\text{Cost}(q) = \mathcal{C}_{\text{in}}(q) + \mathcal{C}_{\text{out}}(q)
\end{equation}


\begin{figure}[htbp]
\centering
\includegraphics[width=0.87\columnwidth]{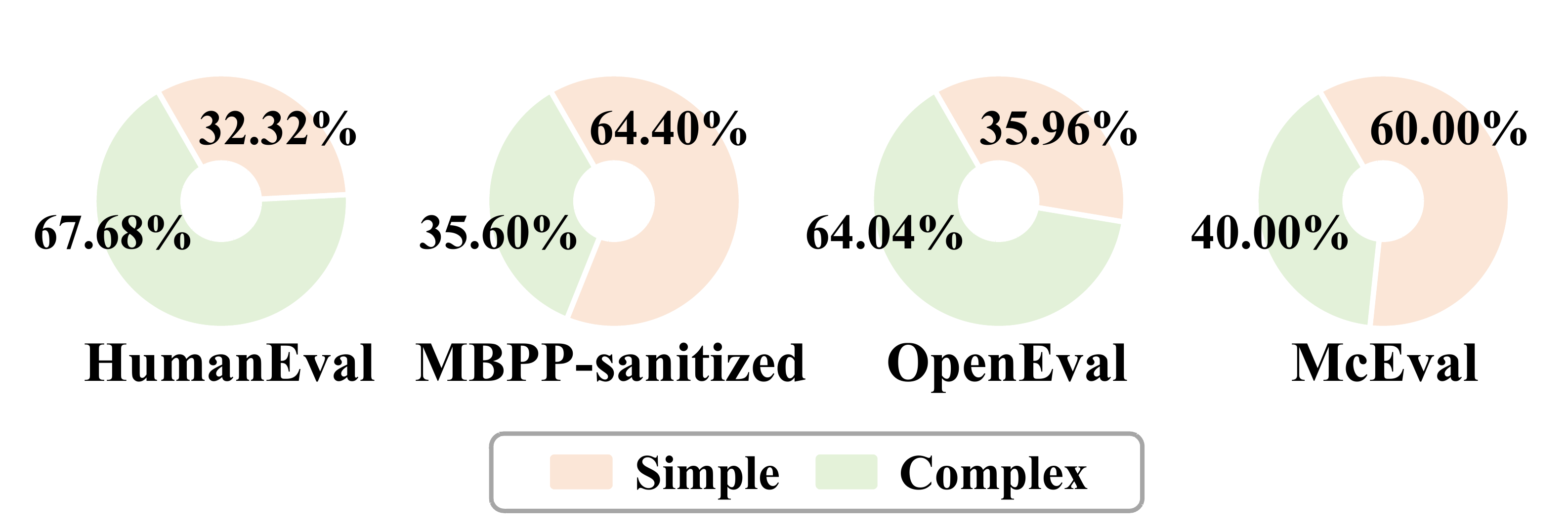} 

\caption{Difficulty perception results across four evaluation benchmarks. These distributions reflect the model’s understanding of task complexity and inform the routing decisions in RoutingGen.}
\label{fig3}
\end{figure}

\subsection{Sampling Settings}
Following recent work~\cite{jiang2024self,li2025structured}, we employ nucleus sampling with top-$p$ filtering to generate candidate programs, ensuring fair comparison across all methods. By default, we generate 20 candidates per problem. For single-stage approaches such as zero-shot and few-shot prompting, we set the sampling temperature to 0.8, top-$p$ to 0.95, and the maximum output length to 300 tokens. In few-shot settings, we select three representative question-code examples, and apply the same configuration in RoutingGen to ensure consistency with the baseline. For Self-CoT, the maximum length is extended to 600 tokens to accommodate longer reasoning chains, while other parameters remain unchanged. For multi-stage approaches including SCoT, ICoT, and Self-Planning, we sample 20 reasoning chains with temperature 0.8, followed by deterministic code generation with temperature 0. Both stages are capped at 300 tokens. An exception is made for DeepSeek-V3, where we generate 5 candidates per problem due to API constraints. All other configurations remain identical to the above.

\section{Experimental Results}

\subsection{Effectiveness and Efficiency of RoutingGen}

\label{sec:effectiveness_RoutingGen}

This section presents a comprehensive evaluation of the proposed RoutingGen framework from three perspectives: overall performance as reported in Table~\ref{tab:result_pass@}, generation cost in terms of token usage in Table~\ref{tab:Token_Consumption}, and difficulty-aware routing outcomes illustrated in Figure~\ref{fig3}.

\paragraph{Main Performance.} As presented in Table~\ref{tab:result_pass@}, RoutingGen achieves state-of-the-art Pass@1 performance across most models and benchmarks, demonstrating both accuracy and generality. 

\begin{table}[htbp]
\centering
\setlength{\tabcolsep}{1mm}
\small
\begin{tabular}{cccccc}
\toprule
\textbf{MT} & \textbf{HE} & \textbf{MP} & \textbf{OE} & \textbf{ME} & \textbf{Avg} \\
\midrule
\textbf{RT} & 42,872 & 61,001 & 35,328 & 0 & 34,800 \\
\midrule
\multicolumn{6}{l}{\textbf{Qwen2.5-Coder-3B-Instruct}} \\
\midrule
SCoT           
& 4,191,205 & 12,857,146 & 4,294,749 
& 1,501,005 & 5,711,026 \\

ICoT          
& 4,275,896 & 12,770,188 & 4,490,805 
& 1,560,068 & 5,774,239 \\

RG     
& 3,309,539 & 4,909,856  & 3,149,950 
& 835,969   & 3,051,329 \\

\cmidrule{2-6} 
R2S 
& \textbf{881,666}   & \textbf{7,947,290} & \textbf{1,144,799} 
& \textbf{665,036}   & \textbf{2,659,698} \\
               
& \textbf{(21.04\%)} & \textbf{(61.81\%)}  & \textbf{(26.66\%)} 
& \textbf{(44.31\%)} & \textbf{(46.57\%)} \\

\cmidrule{2-6}
R2I 
& \textbf{966,357}   & \textbf{7,860,332}  & \textbf{1,340,855} 
& \textbf{724,099}   & \textbf{2,722,911} \\

& \textbf{(22.60\%)} & \textbf{(61.55\%)}  & \textbf{(29.86\%)} 
& \textbf{(46.41\%)} & \textbf{(47.16\%)} \\

\midrule

\multicolumn{6}{l}{\textbf{DeepSeek-Coder-6.7B-Instruct}} \\
\midrule
SCoT           
& 5,118,415 & 14,762,151 & 5,295,607 
& 1,865,235 & 6,760,352 \\

ICoT           
& 4,980,141 & 14,706,320 & 4,867,249 
& 1,760,589 & 6,578,575 \\

RG     
& 3,774,494 & 6,020,323  & 3,429,603 
& 925,268   & 3,537,422 \\

\cmidrule{2-6}
R2S 
& \textbf{1,343,921} & \textbf{8,741,828}  & \textbf{1,866,004 }
& \textbf{939,967}   & \textbf{3,222,930} \\
               
& \textbf{(26.26\%)} & \textbf{(59.22\%)}  & \textbf{(35.24\%)}
& \textbf{(50.39\%)} & \textbf{(47.67\%)} \\

\cmidrule{2-6}

R2I 
& \textbf{1,205,647} & \textbf{8,685,997}  & \textbf{1,437,646} 
& \textbf{835,321}   & \textbf{3,041,153} \\

& \textbf{(24.21\%)} & \textbf{(59.06\%)}  & \textbf{(29.54\%)} 
& \textbf{(47.45\%)} & \textbf{(46.23\%)} \\

\midrule
\multicolumn{6}{l}{\textbf{DeepSeek-V3-671B}} \\
\midrule
SCoT           
& 1,333,100 & 4,335,969  & 1,399,146 
& 471,552   & 1,884,942 \\

ICoT           
& 1,225,360 & 4,071,784  & 1,324,069
& 416,401   & 1,759,404 \\

RG     
&   993,628 & 1,907,516  & 1,050,021 
& 230,027   & 1,045,298 \\

\cmidrule{2-6}
R2S 
& \textbf{339,472} & \textbf{2,428,453}  &   \textbf{349,125} 
& \textbf{241,525} & \textbf{839,644} \\

& \textbf{(25.46\%)} & \textbf{(56.01\%)}  & \textbf{(24.95\%)} 
& \textbf{(51.22\%)} & \textbf{(44.54\%)} \\

\cmidrule{2-6}
R2I 
& \textbf{231,732} & \textbf{2,164,268}  &   \textbf{274,048} 
& \textbf{186,374}   &   \textbf{714,106} \\

& \textbf{(18.91\%)} & \textbf{(53.15\%)}  & \textbf{(20.70\%) }
& \textbf{(44.76\%)} & \textbf{(40.59\%)} \\
\bottomrule
\end{tabular}
\caption{
Total token usage of SCoT, ICoT, and RoutingGen across benchmarks. RT denotes token usage of the routing module; RG indicates the total usage of RoutingGen including inference and routing tokens. R2S and R2I represent RG’s token reduction relative to SCoT and ICoT, showing both absolute and percentage decreases. Avg denotes the average of per-benchmark token reductions and the corresponding relative proportion with respect to the total token usage of the baseline method. MT refers to the prompting method. HE, MP, OE, and ME refer to the HumanEval, MBPP-sanitized, OpenEval, and McEval benchmarks, respectively.
}
\label{tab:Token_Consumption}
\end{table}

\begin{figure*}[htbp]
\centering
\includegraphics[width=2\columnwidth, height=0.165\textheight]{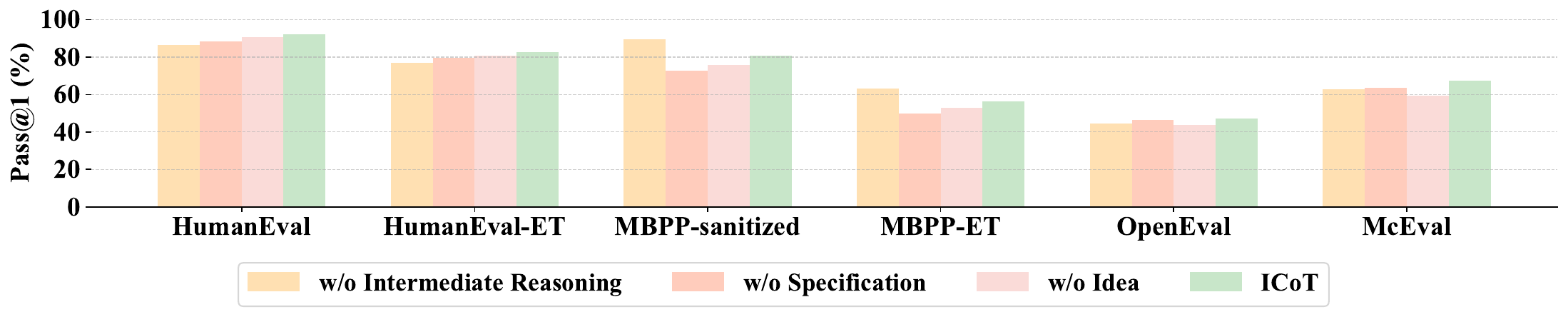} 
\caption{Pass@1 accuracy of ICoT and its ablated variants across six benchmarks using DeepSeek-V3-671B. “ICoT” denotes the model with both specification and idea stages. “w/o Intermediate Reasoning” removes the structured reasoning component entirely. “w/o Specification” omits the specification stage, while “w/o Idea” excludes the idea stage.}
\label{fig4}
\end{figure*}

For instance, it reaches 90.21\% on MBPP-sanitized and 91.83\% on HumanEval with DeepSeek-V3-671B. Notably, structured prompts underperform compared to simpler few-shot approaches on benchmarks such as MBPP-sanitized.  The efficiency of RoutingGen derives from its difficulty-aware routing strategy. As shown in Figure~\ref{fig3}, a majority of tasks in MBPP-sanitized and McEval are classified as \textit{Simple}, accounting for 64.40\% and 60.00\%, respectively, while HumanEval and OpenEval exhibit substantially lower simple-task proportions, both below 36\%. 

As shown in Table~\ref{tab:Token_Consumption}, RoutingGen consistently and substantially reduces token usage compared to both SCoT and ICoT across all evaluated models and benchmarks. All reported token usage values include both routing tokens and inference tokens. On average, RoutingGen achieves a 46.37\% relative reduction in total token usage across settings. For instance, on the MBPP-sanitized benchmark with DeepSeek-Coder-6.7B-Instruct, RoutingGen reduces token usage by 8.74 million compared to SCoT, corresponding to a 59.22\% reduction. Similar trends are consistently observed across other datasets and models, underscoring the efficiency gains introduced by RoutingGen’s difficulty-aware routing strategy.

\textbf{Analysis.} The results empirically validate a key limitation: indiscriminate use of complex prompting strategies often leads to overthinking on simple tasks, resulting in reduced performance and increased computational cost. On MBPP-sanitized and MBPP-ET, which contain a higher proportion of simpler problems, several structured prompting methods perform worse than simpler zero-shot or few-shot baselines. For instance, on MBPP-ET, Self-CoT with DeepSeek-Coder-6.7B-Instruct exhibits the largest performance drop, with its Pass@1 score falling 22.99\% below the zero-shot baseline.

In contrast, RoutingGen mitigates this limitation by dynamically selecting prompting strategies based on task difficulty, achieving more accurate and efficient code generation. On MBPP-sanitized (64.40\% Simple by our classifier), RoutingGen with DeepSeek-Coder-6.7B-Instruct routes most tasks to few-shot generation, achieving a 59.22\% token reduction and an accuracy improvement of 4.60 percentage points over the resource-intensive SCoT (72.30\% vs. 67.70\%). Conversely, on HumanEval (67.68\% Complex by our classifier), RoutingGen achieves a comparable performance (73.51\%) to ICoT (73.14\%), with a moderate token reduction of 24.21\%, reflecting the necessary computational investment for harder problems. The consistent performance across three distinct models and six benchmarks demonstrates the robustness of RoutingGen to variations in model architecture and task distribution. In particular, discrepancies between our baseline results and official reports are expected, as our standardized instructions used for fair cross-model comparison differ from the highly optimized model-specific prompts designed to maximize reported performance~\cite{luo2024wizardcoder}. Detailed experimental settings and results are provided in the Appendix.

\subsection{Effectiveness of ICoT Prompting}

To further assess the standalone effectiveness of ICoT within the RoutingGen framework, we evaluate ICoT under a static prompting configuration without dynamic routing.
 
\textbf{Results.} 
As shown in Table~\ref{tab:result_pass@}, ICoT consistently outperforms baseline prompting methods across the majority of models and benchmarks. The gains are particularly notable when using Qwen2.5-Coder-3B-Instruct and DeepSeek-Coder-6.7B-Instruct. For instance, with DeepSeek-Coder-6.7B-Instruct, ICoT achieves a Pass@1 score of 38.23\% on OpenEval, representing a 128.38\% relative improvement over the zero-shot baseline. Similarly, with Qwen2.5-Coder-3B-Instruct, ICoT demonstrates strong performance across all six benchmarks, including a 45.15\% relative improvement on McEval. Moreover, ICoT demonstrates strong scalability on DeepSeek-V3-671B, achieving a Pass@1 score of 82.68\% on HumanEval-ET and 67.20\% on McEval. These results represent relative improvements of 6.10\% and 102.41\% over the respective zero-shot baselines. 

\textbf{Analysis.} 
The effectiveness of ICoT stems from its two-stage process, which guides the model from explicit modeling of task requirements to core algorithmic design, as illustrated in the “same\_chars” example (Figure~\ref{fig0}(B)). In the Specification stage, the model grounds the task by defining the inputs (two strings s0 and s1) and the functional requirement of the output (a boolean indicating whether both strings contain the same set of characters). Crucially, the subsequent Idea stage abstracts the core algorithmic logic. Instead of prescribing a surface-level procedural loop as in the SCoT baseline, it formulates an intention abstraction: “convert both strings to sets of characters” and then “compare the sets.” This abstraction, which also includes an explicit consideration of time complexity ($O(m+n)$), directly steers code generation toward a concise and correct solution: “return set(s0) == set(s1)”.

\begin{table}[htbp]
\centering
\small
\setlength{\tabcolsep}{1mm}
\begin{tabular}{cccccc}
\toprule
\textbf{Dataset} & \textbf{Method} 
& \textbf{Simple} & \textbf{Complex} 
& \textbf{Total} & \textbf{Pass@1} \\
\midrule

\multirow{3}{*}{HumanEval}   
& Qwen3-8B             
& 53 & 111 & 164 & 73.51\% \\
& GPT-4o              
& 78 & 86  & 164 & 72.99\% \\
\cmidrule{2-6}
& \multicolumn{5}{l}{\textit{Different pairs}: 37 \quad \textit{Different rate}: 22.56\%} \\

\midrule
\multirow{3}{*}{MBPP-sanitized} 

& Qwen3-8B             
& 275 & 152 & 427 & 72.30\% \\
& GPT-4o              
& 363 & 64  & 427 & 72.79\% \\

\cmidrule{2-6}
& \multicolumn{5}{l}{\textit{Different pairs}: 100 \quad \textit{Different rate}: 23.42\%} \\

\midrule

\multirow{3}{*}{OpenEval}
& Qwen3-8B             & 64  & 114 & 178 & 38.74\% \\
& GPT-4o              & 69  & 109 & 178 & 38.51\% \\
\cmidrule{2-6}
& \multicolumn{5}{l}{\textit{Different pairs}: 15 \quad \textit{Different rate}: 8.43\%} \\
\bottomrule
\end{tabular}
\caption{Comparison of difficulty-aware routing and code generation across models. This table compares task-level difficulty labels Simple and Complex classified by Qwen3-8B and GPT-4o, and reports Pass@1 scores based on DeepSeek-Coder-6.7B-Instruct. The “Different pairs” indicates the number of tasks with conflicting difficulty labels between the two models, while “Different rate” denotes their proportion relative to the total.
}
\label{tab:difficulty_comparison}
\end{table}
\subsection{Ablation Analysis}

\paragraph{Robustness to Difficulty Classifier Variants.} We evaluate the robustness of RoutingGen under different difficulty classifiers by comparing Qwen3-8B and GPT-4o on three benchmarks. As shown in Table~\ref{tab:difficulty_comparison}, the two models produce conflicting difficulty labels for 22.56\% of HumanEval, 23.42\% of MBPP-sanitized, and 8.43\% of OpenEval tasks. Despite this variation, RoutingGen consistently outperforms all baseline methods under both classifiers. On MBPP-sanitized, it achieves Pass@1 scores of 72.30\% and 72.79\% with Qwen3-8B and GPT-4o, respectively. Similar trends are also observed when applying self-routing. The consistent gains across classifier variants show that RoutingGen generalizes well under different difficulty estimation conditions.

\paragraph{Effectiveness of ICoT Components.} As shown in Figure~\ref{fig4}, the full ICoT method consistently outperforms all ablated variants on the more challenging benchmarks, confirming the synergistic contribution of both the \textit{Specification} and \textit{Idea} stages. For example, on HumanEval, removing the Specification stage reduces Pass@1 from 92.07\% to 88.17\%, while on McEval, removing the Idea stage leads to a drop from 67.20\% to 59.20\%. Concurrently, the variant without intermediate reasoning achieves the strongest results on the simpler MBPP-sanitized and MBPP-ET datasets. This supports our prior finding that elaborate prompt structures can be counterproductive for simple problems.

\section{Conclusion}
In this work, we address two key limitations in code generation: overthinking caused by uniformly applying structured prompting on simple tasks and the lack of intention abstraction in existing methods. We propose RoutingGen, a difficulty-aware routing framework that dynamically adapts prompting strategies using a classifier to direct simple problems to direct generation and complex ones to ICoT, a two-stage reasoning process that formulates specifications and algorithmic ideas. Experiments on three models and six benchmarks demonstrate that RoutingGen achieves state-of-the-art performance with significantly reduced token usage, and ICoT consistently outperforms other prompting baselines on challenging benchmarks. 

\section{Acknowledgments}
This work was supported in part by the National Natural Science Foundation of China (No. 62372071), the Chongqing Technology Innovation and Application Development Project (No. CSTB2022TIAD-STX0007 and No. CSTB2023TIAD-STX0025), and in part by
the Fundamental Research Funds for the Central Universities under Grant
2023CDJKYJH013.

\bibliography{aaai2026}

\appendix
\onecolumn
\section*{Appendix}

\section{Details of Benchmarks}
Following recent work in LLM evaluation~\cite{openai2024gpt4technicalreport,jiang2024self,yang2025qwen3technicalreport}, we evaluate our methods on six widely adopted code generation benchmarks: 

\textbf{HumanEval} \cite{chen2021evaluatinglargelanguagemodels} is a function-level code generation dataset released by OpenAI, consisting of 164 manually written Python problems. Each problem includes a function signature, a natural language description, a reference implementation, and a set of unit tests (approximately 7.7 per problem) for verifying functional correctness.

\textbf{MBPP-sanitized}\cite{austin2021programsynthesislargelanguage} is a manually verified subset of the MBPP (Mostly Basic Python Problems) dataset, originally constructed via crowdsourcing by Google. It contains 427 function-level Python programming tasks, covering fundamental programming skills and typical usage of standard library functions. Each problem consists of a natural language description, a reference code implementation, and three automated test cases for validating functional correctness.

\textbf{HumanEval-ET and MBPP-ET}\cite{dong2025codescore} are two publicly available extended versions of MBPP and HumanEval, respectively, each augmenting every task with over 100 additional test cases. By incorporating a broader range of edge cases, these extended benchmarks significantly enhance the completeness and robustness of code evaluation compared to their original counterparts.

\textbf{OpenEval}\cite{yang2024chain}is a code generation benchmark constructed from the AVATAR code translation dataset, comprising 178 competition-level programming problems. Each problem includes a natural language description, a reference implementation, and five manually designed functional test cases.

\textbf{McEval}\cite{chai2024mceval}is a multilingual benchmark for code generation. In this work, we use its Python subset, which consists of 50 function-level problems. Each problem includes a natural language description, a function signature, a reference implementation, and a set of test cases, covering fundamental programming skills such as mathematical computation and control flow.

\section{The RoutingGen Framework}

\begin{algorithm}[ht]
\caption{The RoutingGen Framework}
\label{alg:routinggen1}

\begin{algorithmic}

\STATE \textbf{Input:} Input problem $q$; Classifier model $\mathcal{M}_{\text{cls}}$; Generator model $\mathcal{M}_{\text{gen}}$; Number of samples $n$; Prompt templates $T_{\text{cls}}, T_{\text{Direct}}, T_{\text{ICoT}}^{(1)}, T_{\text{ICoT}}^{(2)}$.
\STATE \textbf{Output:} A set of $n$ candidate code solutions $\mathcal{C}$.

\STATE
\STATE $(d^*, r^*) = \underset{(d, r) \text{ s.t. } d \in \mathcal{L}}{\arg\max} \, P_{\mathcal{M}_{\text{cls}}}(d, r \mid q, T_{\text{cls}})$
\hspace{0.8em}\text{// Difficulty-Aware Dynamic Routing}

\STATE

\IF{$d^* == \text{\textit{Simple}}$}
    \STATE $C_i^* \sim P_{\mathcal{M}_{\text{gen}}}(\cdot \mid q, T_{\text{Direct}})$ for $i = 1, \dots, n$ \hspace{0.8em}\text{// Few-shot Generation}
    \STATE $\mathcal{C} \leftarrow \{C_1^*, \dots, C_n^*\}$
    \RETURN $\mathcal{C}$
\ELSE
    \STATE $\text{ICoT}_i \sim P_{\mathcal{M}_{\text{gen}}}(\cdot \mid q, T_{\text{ICoT}}^{(1)})$ for $i = 1, \dots, n$ \hspace{0.8em}\text{// ICoT Generation}
    \STATE $\mathcal{R}_{\text{ICoT}} \leftarrow \{\text{ICoT}_1, \dots, \text{ICoT}_n\}$

    \STATE

    \FOR{$i = 1$ \TO $n$}
        \STATE $C_i^* \leftarrow \texttt{GreedyDecode}_{\mathcal{M}_{\text{gen}}}(q, \text{ICoT}_i, T_{\text{ICoT}}^{(2)})$ \hspace{1em}\text{// Code Generation}
        \STATE $\mathcal{C} \leftarrow \mathcal{C} \cup \{C_i^*\}$
    \ENDFOR
    \RETURN $\mathcal{C}$
\ENDIF

\end{algorithmic}
\end{algorithm}

\newpage
\section{Self-Routing: Difficulty-Aware Routing without External Classifiers}

\begin{table*}[ht]
\centering
\small
\setlength{\tabcolsep}{1mm}
\resizebox{\textwidth}{!}{%
\begin{tabular}{@{}ccccccc@{}}
\toprule
 \textbf{Method} & \textbf{HumanEval} & \textbf{HumanEval-ET} & \textbf{MBPP-sanitized} & \textbf{MBPP-ET} & \textbf{OpenEval} & \textbf{McEval} \\
\midrule

\multicolumn{7}{l}{\textbf{Qwen2.5-Coder-3B-Instruct}} \\ 
\midrule






RG-Self
& \textbf{76.62\%} \textbf{(+1.50\%)} & \textbf{69.36\%} \textbf{(+3.08\%)} 
& \underline{68.70\% (+11.85\%)} & \textbf{48.28\%} \textbf{(+9.43\%)}
& \textbf{35.87\%} \textbf{(+2.31\%)} & \textbf{35.30\%} \textbf{(+31.72\%)} \\ 

ICoT 
& \textbf{77.10\%} \textbf{(+2.13\%)} & \textbf{69.73\%} 
\textbf{(+3.63\%)} & \textbf{69.11\%} \textbf{(+12.52\%)} 
& \textbf{48.58\%} \textbf{(+10.11\%)} & \textbf{35.70\%} 
\textbf{(+1.83\%)} & \textbf{38.90\%} \textbf{(+45.15\%)} \\ 

\midrule
\multicolumn{7}{l}{\textbf{DeepSeekCoder-6.7B-Instruction}} \\ 
\midrule






RG-Self
& \textbf{73.11\%} \textbf{(+59.11\%)} & \textbf{64.39\%} \textbf{(+62.60\%)} 
& \underline{72.60\% (+57.55\%)} & \textbf{51.86\%} \textbf{(+66.75\%)}
& \underline{37.56\% (+124.37\%)} & \textbf{41.00\%} \textbf{(+72.27\%)} \\ 

ICoT 
& \textbf{73.14\%} \textbf{(+59.17\%)} & \textbf{65.03\%} \textbf{(+64.22\%)}
& \underline{71.24\% (+54.60\%)} & \underline{50.85\% (+63.50\%)} 
& \textbf{38.23\%} \textbf{(+128.38\%)} & \textbf{42.90\%} \textbf{(+80.25\%)} \\ 

\midrule
\multicolumn{7}{l}{\textbf{DeepSeek-V3-671B}} \\ 
\midrule







RG-Self
& \textbf{91.59\%} \textbf{(+6.99\%)} & \underline{81.95\% (+5.16\%)} 
& \textbf{90.12\%} \textbf{(+1.27\%)} & \textbf{63.33\% \textbf(+0.46\%)} 
& \textbf{48.99\%} \textbf{(+4.30\%)} & \textbf{65.20\%} \textbf{(+96.39\%)} \\

ICoT 
& \textbf{92.07\%} \textbf{(+7.55\%)} & \textbf{82.68\%} \textbf{(+6.10\%)} 
& 80.70\% (-9.31\%) & 56.25\% (-10.77\%) & \textbf{47.30\%} \textbf{(+0.70\%)}
& \textbf{67.20\%} \textbf{(+102.41\%)} \\

\bottomrule
\end{tabular}
}
\caption{
Pass@1 comparison under the Self-Routing setting across three models and six code generation benchmarks.
}
\label{tab:self-routing}
\end{table*}

\paragraph{Results and Analysis.} We report the results in Table~\ref{tab:self-routing} under the self-routing setting, where each model autonomously performs difficulty-aware routing, eliminating any reliance on an external difficulty classifier. The table presents Pass@1 results across three models and six benchmarks. Despite differences in routing decisions, the resulting generation performance remains broadly comparable between self-routing and routing with external classifiers. Overall, RG-Self achieves state-of-the-art performance in the majority of settings while maintaining competitive results in the remaining cases. For instance, Qwen2.5-Coder-3B-Instruct attains 69.36\% on HE-ET and 48.28\% on MP-ET, while DeepSeekCoder-6.7B-Instruction achieves 64.39\% and 51.86\%, respectively, consistently surpassing all baselines.

\section{HumanEval-X Evaluation on C++ Code Generation}

\begin{table*}[ht]
\centering
\small
\setlength{\tabcolsep}{1mm}
\resizebox{0.7\textwidth}{!}{%
\begin{tabular}{@{}ccccccccc@{}}
\toprule
 \textbf{zero-shot} & \textbf{few-shot} & \textbf{Self-CoT} & \textbf{ZS-CoT}  & \textbf{SP} & \textbf{SCoT} & \textbf{few-shot-CoT} & \textbf{RG} & \textbf{ICoT} \\
\midrule

53.66\% & 81.95\% & 42.44\%
& 54.39\% & 73.54\% & 78.66
& 77.56\% & \textbf{82.44\%} & \underline{81.46\%} \\

\bottomrule
\end{tabular}
}
\caption{
Pass@1 performance on HumanEval-X for C++ code generation.
}
\label{tab:humaneval-x}
\end{table*}

\paragraph{Results and Analysis.} Table~\ref{tab:humaneval-x} reports the Pass@1 results on HumanEval-X \cite{zheng2023codegeex}, a multilingual extension of HumanEval that evaluates functional correctness across multiple programming languages; in this evaluation, we focus on the C++ subset. Few-shot-CoT uses the same number of exemplars as SCoT and ICoT, but differs in the form of chain-of-thought employed, allowing comparison across distinct reasoning paradigms. Overall, RoutingGen and ICoT consistently outperform direct generation and structured prompting baselines, demonstrating their effectiveness beyond Python-only benchmarks. Notably, in this setting RoutingGen adopts self-routing and is evaluated with DeepSeek-V3-671B, while still achieving strong and stable performance.

\section{LiveCodeBench Evaluation}
\begin{table*}[ht]
\centering
\small
\setlength{\tabcolsep}{1mm}
\resizebox{0.7\textwidth}{!}{%
\begin{tabular}{@{}ccccccccc@{}}
\toprule
 \textbf{zero-shot} & \textbf{few-shot} & \textbf{Self-CoT} & \textbf{ZS-CoT}  & \textbf{SP} & \textbf{SCoT} & \textbf{few-shot-CoT} & \textbf{RG} & \textbf{ICoT} \\
\midrule

33.00\% & 42.55\% & 29.95\%
& 29.10\% & 37.20\% & 43.35
& 40.55\% & \underline{44.50\%} & \textbf{45.10\%} \\

\bottomrule
\end{tabular}
}
\caption{
Pass@1 performance on LiveCodeBench.
}
\label{tab:LiveCodeBench}
\end{table*}

\paragraph{Results and Analysis.} Table~\ref{tab:LiveCodeBench} reports the Pass@1 results on LiveCodeBench \cite{jain2024livecodebenchholisticcontaminationfree}, a comprehensive and contamination-free benchmark for evaluating LLMs on code that continuously collects new problems from competitive programming platforms and covers a broad range of code-related capabilities beyond code generation. Overall, RoutingGen and ICoT achieve the strongest performance among all compared methods, outperforming both direct generation baselines and structured prompting approaches. In particular, RoutingGen achieves 44.50\% Pass@1, while the ICoT-based setting attains 45.10\%, with both results outperforming the zero-shot, few-shot, and CoT-based baselines. Notably, in this evaluation RoutingGen leverages the dataset-provided labels for routing decisions, yet still maintains stable and competitive performance.

\newpage
\section{Comparative Analysis of Difficulty Classifiers}

We evaluate the impact of different difficulty classifiers by comparing Pass@1 accuracy and token usage under Qwen3-8B and GPT-4o routing.

\subsection{Pass@1 Performance under Qwen3-8B and GPT-4o Routing}

\begin{table*}[ht]
\centering
\small
\setlength{\tabcolsep}{1mm}
\resizebox{\textwidth}{!}{%
\begin{tabular}{@{}ccccccc@{}}
\toprule
 \textbf{Method} & \textbf{HumanEval} & \textbf{HumanEval-ET} & \textbf{MBPP-sanitized} & \textbf{MBPP-ET} & \textbf{OpenEval} & \textbf{McEval} \\
\midrule

\multicolumn{7}{l}{\textbf{Qwen2.5-Coder-3B-Instruct}} \\ 
\midrule
zero-shot 
& 75.49\% & 67.29\% & 61.42\%
& 44.12\% & 35.06\% & 26.80\% \\

few-shot 
& 72.80\% (-3.56\%) & 65.91\% (-2.05\%) 
& \underline{68.74\% (+11.92\%)} & \underline{48.20\% (+9.25\%)} 
& 34.75\% (-0.88\%) & 31.10\% (+16.04\%) \\

Self-CoT 
& 71.55\% (-5.22\%) & 64.18\% (-4.62\%) 
& 66.01\% (+7.47\%) & 46.51\% (+5.42\%) 
& 34.13\% (-2.65\%) & 23.50\% (-12.31\%) \\

ZS-CoT
& \underline{75.70\% (+0.28\%)} & \underline{67.99\% (+1.04\%)} 
& 66.73\% (+8.65\%) & 47.24\% (+7.07\%) 
& 35.42\% (+1.03\%) & 26.20\% (-2.24\%) \\

SP
& 72.84\% (-3.51\%) & 64.02\% (-4.86\%) 
& 53.69\% (-12.59\%) & 36.93\% (-16.30\%) 
& \underline{35.65\% (+1.68\%)} & 25.30\% (-5.60\%) \\

SCoT
& 65.27\% (-13.54\%) & 58.60\% (-12.91\%) 
& 61.62\% (+0.33\%) & 41.87\% (-5.10\%) 
& 33.57\% (-4.25\%) & \underline{35.10\% (+30.97\%)} \\
\cmidrule{2-7} 

RG-Qwen
& \textbf{76.65\%} \textbf{(+1.54\%)} & \textbf{69.02\%} \textbf{(+2.57\%)} 
& \textbf{68.84\%} \textbf{(+13.71\%)} & \textbf{49.33\%} \textbf{(+11.81\%)}
& \textbf{35.76\%} \textbf{(+2.00\%)} & \textbf{35.30\%} \textbf{(+31.72\%)} \\ 

RG-GPT
& \textbf{77.90\%} \textbf{(+3.19\%)} & \textbf{70.82\%} \textbf{(+5.25\%)} 
& \textbf{68.79\%} \textbf{(+12.00\%)} & \textbf{48.59\%} \textbf{(+10.13\%)}
& \textbf{36.21\%} \textbf{(+3.28\%)} & \textbf{35.30\%} \textbf{(+31.72\%)} \\

ICoT 
& \textbf{77.10\%} \textbf{(+2.13\%)} & \textbf{69.73\%} 
\textbf{(+3.63\%)} & \textbf{69.11\%} \textbf{(+12.52\%)} 
& \textbf{48.58\%} \textbf{(+10.11\%)} & \textbf{35.70\%} 
\textbf{(+1.83\%)} & \textbf{38.90\%} \textbf{(+45.15\%)} \\ 

\midrule
\multicolumn{7}{l}{\textbf{DeepSeekCoder-6.7B-Instruction}} \\ 
\midrule
zero-shot 
& 45.95\% & 39.60\% & 46.08\% 
& 31.10\% & 16.74\% & 23.80\% \\

few-shot 
& \underline{72.68\% (+58.17\%)} & \underline{64.27\% (+62.30\%)}
& \textbf{73.40\% (+59.29\%)} & \underline{51.64\% (+66.05\%)} 
& \underline{37.89\% (+126.34\%)} & 39.30\% (+65.13\%) \\

Self-CoT
& 68.29\% (+48.62\%) & 60.27\% (+52.20\%)
& 36.70\% (-20.36\%) & 23.95\% (-22.99\%) 
& 35.34\% (+111.11\%) & 38.70\% (+62.61\%) \\

ZS-CoT
& 63.66\% (+38.54\%) & 55.58\% (+40.35\%) 
& 38.01\% (-17.51\%) & 24.75\% (-20.42\%) 
& 34.35\% (+105.20\%) & 34.80\% (+46.22\%) \\

SP 
& 58.90\% (+28.18\%) & 52.47\% (+32.50\%) 
& 49.19\% (+6.75\%) & 33.72\% (+8.42\%)
& 25.17\% (+50.36\%) & 30.50\% (+28.15\%) \\

SCoT 
& 70.79\% (+54.06\%) & 62.87\% (+58.76\%) 
& 67.70\% (+46.92\%) & 46.58\% (+49.77\%) 
& 36.91\% (+120.49\%) & \underline{40.90\% (+71.85\%)} \\
\cmidrule{2-7} 

RG-Qwen
& \textbf{73.51\%} \textbf{(+59.97\%)} & \textbf{64.76\%} \textbf{(+63.53\%)} 
& \underline{72.30\% (+56.90\%)} & \textbf{51.73\%} \textbf{(+66.33\%)}
& \textbf{38.74\%} \textbf{(+131.42\%)} & \textbf{41.00\%} \textbf{(+72.27\%)} \\ 

RG-GPT
& \textbf{72.99\%} \textbf{(+58.85\%)} & \textbf{64.30\%} \textbf{(+62.37\%)} 
& \underline{72.79\% (+57.96\%)} & \textbf{53.45\%} \textbf{(+71.86\%)}
& \textbf{38.51\%} \textbf{(+130.05\%)} & \textbf{41.00\%} \textbf{(+72.27\%)} \\

ICoT 
& \textbf{73.14\%} \textbf{(+59.17\%)} & \textbf{65.03\%} \textbf{(+64.22\%)}
& \underline{71.24\% (+54.60\%)} & \underline{50.85\% (+63.50\%)} 
& \textbf{38.23\%} \textbf{(+128.38\%)} & \textbf{42.90\%} \textbf{(+80.25\%)} \\ 

\midrule
\multicolumn{7}{l}{\textbf{DeepSeek-V3-671B}} \\ 
\midrule

zero-shot 
& 85.61\% & 77.93\% & {88.99\%} 
& 63.04\% & 46.97\% & 33.20\% \\

few-shot 
& 84.76\% (-0.99\%) & 75.73\% (-2.82\%) 
& \underline{89.23\% (+0.27\%)} & \textbf{63.23\% (+0.30\%)} 
& 42.81\% (-8.86\%) & 51.60\% (+55.42\%) \\

Self-CoT 
& \underline{91.34\% (+6.69\%)} & 81.71\% (+4.85\%) 
& 83.98\% (-5.63\%) & 60.09\% (-4.68\%) 
& 45.51\% (-3.11\%) & \underline{64.40\% (+93.98\%)} \\

ZS-CoT
& 90.85\% (+6.12\%) & \underline{82.32\% (+5.63\%)}
& 84.92\% (-4.57\%) & 60.42\% (-4.16\%) 
& 41.01\% (-12.69\%) & 54.00\% (+62.65\%) \\

SP 
& 80.37\% (-6.12\%) & 73.17\% (-6.11\%) 
& 80.52\% (-9.52\%) & 55.93\% (-11.28\%) 
& 42.81\% (-8.86\%) & 51.60\% (+55.42\%) \\

SCoT 
& 90.98\% (+6.27\%) & 81.34\% (+4.38\%) 
& 79.30\% (-10.89\%) & 54.71\% (-13.21\%)
& \underline{47.08\% (+0.23\%)} & 60.00\% (+80.72\%) \\
\cmidrule{2-7} 

RG-Qwen
& \textbf{91.83\%} \textbf{(+7.27\%)} & \underline{82.07\% (+5.31\%)} 
& \textbf{90.21\%} \textbf{(+1.37\%)} & \underline{63.14\% (+0.16\%)} 
& \textbf{47.98\%} \textbf{(+2.15\%)} & \textbf{65.20\%} \textbf{(+96.39\%)} \\

RG-GPT
& \textbf{91.71\%} \textbf{(+7.13\%)} & \underline{81.71\% (+4.85\%)} 
& \textbf{90.21\%} \textbf{(+1.37\%)} & \underline{63.09\% (+0.08\%)}
& \textbf{47.42\%} \textbf{(+0.96\%)} & \textbf{65.20\%} \textbf{(+96.39\%)} \\

ICoT 
& \textbf{92.07\%} \textbf{(+7.55\%)} & \textbf{82.68\%} \textbf{(+6.10\%)} 
& 80.70\% (-9.31\%) & 56.25\% (-10.77\%) & \textbf{47.30\%} \textbf{(+0.70\%)}
& \textbf{67.20\%} \textbf{(+102.41\%)} \\

\bottomrule
\end{tabular}
}
\caption{
Pass@1 comparison across three models and six benchmarks. 
Results include two direct generation baselines, four structured prompting baselines, and the proposed RoutingGen and ICoT. 
RG-Qwen and RG-GPT denote RoutingGen using Qwen3-8B and GPT-4o as the difficulty classifier, respectively. 
Bold and underline indicate the best and second-best performance among all baseline prompting methods and the two proposed approaches (ICoT and RoutingGen). 
Relative improvements over the zero-shot baseline are reported in parentheses.
}
\label{tab:result_Qwen-GPT_compare_pass@}
\end{table*}

\paragraph{Results and Analysis.} To further assess the stability of RoutingGen under different difficulty classifiers, we present a complementary analysis in Table~\ref{tab:result_Qwen-GPT_compare_pass@}, comparing Qwen3-8B and GPT-4o across three representative benchmarks. The table reports the Pass@1 results of RoutingGen when using Qwen3-8B (denoted as RG-Qwen) and GPT-4o (denoted as RG-GPT) as difficulty classifiers, evaluated across three models and six benchmarks. Although the difficulty labels differ on a portion of problems (22.56\% on HumanEval, 23.42\% on MBPP-sanitized, and 8.43\% on OpenEval), the resulting generation performance remains comparable. For example, on MBPP-ET, RG-Qwen achieves 51.73\% and 63.14\% under DeepSeek-Coder-6.7B-Instruction and DeepSeek-V3-671B, respectively, while RG-GPT attains 53.45\% and 63.09\% on the same models. Consistent trends across all benchmarks, with both variants delivering comparable performance, demonstrate that RoutingGen generalizes well under different difficulty classifiers. Evaluations over three models and six benchmarks further confirm that it achieves state-of-the-art performance in the majority of settings while maintaining competitive results in the remaining cases.

\newpage
\subsection{Token Usage under Qwen3-8B and GPT-4o Routing}

\begin{table*}[ht]
\centering
\setlength{\tabcolsep}{1mm}
\small
\resizebox{0.95\textwidth}{!}{%
\begin{tabular}{ccccccccccc}
\toprule
\multirow{2}{*}{\textbf{Method}} 
& \multicolumn{2}{c}{\textbf{HumanEval}} 
& \multicolumn{2}{c}{\textbf{MBPP-sanitized}} 
& \multicolumn{2}{c}{\textbf{OpenEval}} 
& \multicolumn{2}{c}{\textbf{MCEval}}
& \multicolumn{2}{c}{\textbf{Average}}\\

& \textbf{Input} & \textbf{Output} 
& \textbf{Input} & \textbf{Output} 
& \textbf{Input} & \textbf{Output} 
& \textbf{Input} & \textbf{Output}
& \textbf{Input} & \textbf{Output} \\

\midrule
\rowcolor[gray]{0.88}
\multicolumn{11}{l}{\textbf{\textit{Difficulty Classifier: Qwen3-8B
}}} \\ 
\midrule
Routing       
&38,741  &4,131  
&52,362  &8,639  
&31,084  &4,244  
& - & -
&30,547  &4,254   \\

\midrule
\multicolumn{9}{l}{\textbf{Qwen2.5-Coder-3B-Instruct}} \\ 
\midrule

ICoT          
& 3,097,554 & 1,178,342 
& 10,135,020 & 2,635,168 
& 3,184,859 & 1,305,946 
& 1,093,775 & 466,293
& 4,377,802 & 1,396,437\\

RG-Qwen  
&2,209,529  &1,100,010    
&3,932,478  &977,378  
&2,139,438  &1,010,512 
&460,863  &375,106 
&2,185,577  &865,752 \\

\multirow{2}{*}{Reduction}           
& \textbf{{↓888,025}}  &  \textbf{{↓78,332}} 
& \textbf{{↓6,202,542}}  &  \textbf{{↓1,657,790}} 
& \textbf{{↓1,045,421}}  &  \textbf{{↓295,434}} 
& \textbf{{↓632,912}}  &  \textbf{{↓91,187}} 
& \textbf{{↓2,192,225}}  &  \textbf{{↓530,686}}\\

& \textbf{{(28.67\%)}} &   \textbf{{(6.65\%)}} 
& \textbf{{(61.20\%)}} &   \textbf{{(62.91\%)}} 
& \textbf{{(32.82\%)}} &   \textbf{{(22.62\%)}} 
& \textbf{{(57.86\%)}} &   \textbf{{(19.56\%)}} 
& \textbf{{(50.08\%)}} &   \textbf{{(38.00\%)}}\\

\midrule
\multicolumn{9}{l}{\textbf{DeepSeekCoder-6.7B-Instruction}} \\ 
\midrule

ICoT          
& 3,627,843 & 1,352,298
& 11,603,251 & 3,103,069 
& 3,717,466 & 1,149,783
& 1,263,358 & 497,231 
& 5,052,980  & 1,525,595\\

RG-Qwen     
&2,568,638  &1,205,856    
&4,544,592  &1,475,731  
&2,477,101  &952,502 
&536,659  &388,609 
&2,531,748  &1,005,675 \\

\multirow{2}{*}{Reduction}         
& \textbf{{↓1,059,205}}  &  \textbf{{↓146,442}} 
& \textbf{{↓7,058,659}}  &  \textbf{{↓1,627,338}} 
& \textbf{{↓1,240,365}}  &  \textbf{{↓197,281}} 
& \textbf{{↓726,699}}  &  \textbf{{↓108,622}} 
& \textbf{{↓2,521,232}}  &  \textbf{{↓519,921}}\\

& \textbf{{(29.20\%)}} &   \textbf{{(10.83\%)}} 
& \textbf{{(60.83\%)}} &   \textbf{{(52.44\%)}} 
& \textbf{{(33.37\%)}} &   \textbf{{(17.16\%)}} 
& \textbf{{(57.52\%)}} &   \textbf{{(21.85\%)}} 
& \textbf{{(49.90\%)}} &   \textbf{{(34.08\%)}}\\

\midrule
\multicolumn{9}{l}{\textbf{DeepSeek-V3-671B}} \\ 
\midrule

ICoT         
& 878,920        & 346,440    
& 2,991,048         & 1,080,736  
& 902,375        & 421,694       
& 289,991        & 126,410      
& 1,265,584 & 493,820 \\

RG-Qwen     
&677,253  &316,375    
&1,289,913  &617,603  
&655,561  &394,460 
&141,368  &88,659 
&691,024  &354,274 \\

\multirow{2}{*}{Reduction}          
& \textbf{{↓201,667}}  &  \textbf{{↓30,065}} 
& \textbf{{↓1,701,135}}  &  \textbf{{↓463,133}} 
& \textbf{{↓246,814}}  &  \textbf{{↓27,234}} 
& \textbf{{↓148,623}}  &  \textbf{{↓37,751}} 
& \textbf{{↓574,560}}  &  \textbf{{↓139,546}}\\

& \textbf{{(22.94\%)}} &   \textbf{{(8.68\%)}} 
& \textbf{{(56.87\%)}} &   \textbf{{(42.85\%)}} 
& \textbf{{(27.35\%)}} &   \textbf{{(6.46\%)}} 
& \textbf{{(51.25\%)}} &   \textbf{{(29.86\%)}} 
& \textbf{{(45.40\%)}} &   \textbf{{(28.26\%)}}\\

\midrule
\rowcolor[gray]{0.88}
\multicolumn{11}{l}{\textbf{\textit{Difficulty Classifier: GPT-4o}}} \\ 
\midrule
Routing       
& 38,288 & 6,590 
& 52,332 & 14,764
& 31,046 & 7,119 
& - & -
& 30,417 & 7,118 \\

\midrule
\multicolumn{9}{l}{\textbf{Qwen2.5-Coder-3B-Instruct}} \\ 
\midrule

ICoT          
& 3,097,554 & 1,178,342 
& 10,135,020 & 2,635,168 
& 3,184,859 & 1,305,946 
& 1,093,775 & 466,293
& 4,377,802 & 1,396,437\\

RG-GPT 
& 1,739,915 &   646,201
&  1,894,551 & 1,448,959 
& 2,029,304 & 755,572 
&   460,863 & 375,106 
& 1,531,158 & 806,460\\

\multirow{2}{*}{Reduction}           
& \textbf{{↓1,357,639}} &  \textbf{{↓532,141}} 
&  \textbf{{↓8,240,469}} &   \textbf{{↓1,186,209}} 
& \textbf{{↓1,155,555}} &   \textbf{{↓550,374}} 
&   \textbf{{↓632,912}} &  \textbf{{↓91,187}} 
& \textbf{{↓2,846,644}}  & \textbf{{↓589,978}}\\

& \textbf{{(43.83\%)}} &  \textbf{{(45.16\%)}} 
&  \textbf{{(81.31\%)}} &   \textbf{{(45.01\%)}} 
& \textbf{{(36.28\%)}} &   \textbf{{(42.14\%)}} 
&   \textbf{{(57.86\%)}} &  \textbf{{(19.56\%)}} 
& \textbf{{(65.02\%)}}  & \textbf{{(42.25\%)}}\\

\midrule
\multicolumn{9}{l}{\textbf{DeepSeekCoder-6.7B-Instruction}} \\ 
\midrule

ICoT          
& 3,627,843 & 1,352,298
& 11,603,251 & 3,103,069 
& 3,717,466 & 1,149,783
& 1,263,358 & 497,231 
& 5,052,980  & 1,525,595\\

RG-GPT     
& 2,036,006 & 1,141,565 
&  1,279,756 & 1,056,154
& 2,375,343 &  934,620
&   536,659 & 388,609 
& 1,556,941  & 880,237 \\

\multirow{2}{*}{Reduction}         
& \textbf{{↓1,591,837}} &   \textbf{{↓210,733}} 
& \textbf{{↓10,323,495}} & \textbf{{↓2,046,915}} 
& \textbf{{↓1,342,123}} & \textbf{{↓215,163}}
&   \textbf{{↓726,699}} & \textbf{{↓108,622}} 
& \textbf{{↓3,496,039}} & \textbf{{↓645,358}} \\

& \textbf{{(43.88\%)}} &  \textbf{{(15.58\%)}} 
&  \textbf{{(88.97\%)}} &   \textbf{{(65.96\%)}} 
& \textbf{{(36.10\%)}} &   \textbf{{(18.71\%)}} 
&   \textbf{{(57.52\%)}} &  \textbf{{(21.85\%)}} 
& \textbf{{(69.19\%)}}  & \textbf{{(42.30\%)}}\\

\midrule
\multicolumn{9}{l}{\textbf{DeepSeek-V3-671B}} \\ 
\midrule

ICoT         
& 878,920        & 346,440    
& 2,991,048         & 1,080,736  
& 902,375        & 421,694       
& 289,991        & 126,410      
& 1,265,584 & 493,820 \\

RG-GPT   
& 549,574        & 272,421 
& 732,480         & 537,185    
& 628,980        & 385,313        
& 141,368        & 88,659      
& 513,101        & 320,895 \\

\multirow{2}{*}{Reduction}          
& \textbf{{↓329,346}} &   \textbf{{↓74,019}} 
& \textbf{{↓2,258,568}} & \textbf{{↓543,551}}
& \textbf{{↓273,395}} & \textbf{{↓36,381}}   
& \textbf{{↓148,623}} & \textbf{{↓37,751}} 
& \textbf{{↓752,483}} & \textbf{{↓172,926}}\\

& \textbf{{(37.47\%)}} &  \textbf{{(21.37\%)}} 
&  \textbf{{(75.51\%)}} &   \textbf{{(50.29\%)}} 
& \textbf{{(30.30\%)}} &   \textbf{{(8.63\%)}} 
&   \textbf{{(51.25\%)}} &  \textbf{{(29.86\%)}} 
& \textbf{{(59.46\%)}}  & \textbf{{(35.02\%)}}\\

\bottomrule
\end{tabular}
}
\caption{Comparison of input and output token usage between ICoT and RoutingGen under two difficulty classifiers (Qwen3-8B and GPT-4o) across five benchmarks. “ICoT” refers to the baseline prompting strategy, while “RG-Qwen” and “RG-GPT” denote RoutingGen using Qwen3-8B and GPT-4o as the difficulty classifiers, respectively. All RoutingGen values include both inference tokens and routing overhead. “Reduction” indicates the absolute and relative token savings of RoutingGen compared to ICoT, with percentages reported in parentheses.
}
\label{tab:Token_Consumption_compare}
\end{table*}

\paragraph{Results and Analysis.}
We examine the token usage of RoutingGen under different difficulty classifiers. As shown in Table~\ref{tab:Token_Consumption_compare}, GPT-4o, a larger model than Qwen3-8B, tends to classify a greater proportion of tasks as simple~\cite{openai2024gpt4technicalreport}. Compared to ICoT, RoutingGen with GPT-4o achieves average reductions of 66\% in input tokens and 41\% in output tokens, whereas the Qwen3-8B variant yields corresponding reductions of 49\% and 35\%. While the routing behavior varies due to differences in classifier predictions, both configurations maintain strong Pass@1 performance, as shown in Table~\ref{tab:result_Qwen-GPT_compare_pass@}, and achieve substantial reductions in token consumption across the evaluated benchmarks.

\newpage
\section{Pass@1 Performance by Difficulty under Qwen3-8B and GPT-4o Routing}
\subsection{Simple Subset}
\begin{table*}[htbp]
\centering
\small
\setlength{\tabcolsep}{1.5mm}
\begin{tabular}{cccccccccccc}
\toprule
\multirow{2}{*}{\textbf{Method}} & \multicolumn{2}{c}{\textbf{HumanEval}} & \multicolumn{2}{c}{\textbf{HumanEval-ET}} & \multicolumn{2}{c}{\textbf{MBPP-sanitized}} & \multicolumn{2}{c}{\textbf{MBPP-ET}} & \multicolumn{2}{c}{\textbf{OpenEval}} & \textbf{McEval} \\
\cmidrule{2-12}
& Qwen & GPT & Qwen & GPT & Qwen & GPT & Qwen & GPT & Qwen & GPT & - \\
\midrule
\multicolumn{12}{l}{\textbf{Qwen2.5-Coder-3B-Instruct}} \\
\midrule
zero-shot & 83.77\% & 78.14\% & 75.66\% & 71.86\% & 73.60\% & 66.75\% & 54.16\% & 48.44\% & 50.86\% & 51.59\% & 36.17\% \\

few-shot & \underline{90.85\%} & \underline{82.56\%} & \underline{80.47\%} & \textbf{75.26\%} & \textbf{77.31\%} & \underline{72.74\%} & \textbf{55.45\%} & \textbf{51.86\%} & \textbf{53.98\%} & {52.17\%} & {43.33\%} \\

\cmidrule{2-12}

Self-CoT & 86.89\% & 78.91\% & 76.23\% & 70.90\% & 75.07\% & 70.39\% & 53.91\% & 50.22\% & 53.05\% & 52.10\% & 32.33\% \\

ZS-CoT & \underline{90.85\%} & \textbf{83.27\%} & 79.43\% & \underline{74.94\%} & 75.87\% & 71.20\% & \underline{55.13\%} & 50.99\% & 52.50\% & 52.61\% & 34.00\% \\

SP & 88.77\% & 78.97\% & 78.58\% & 71.22\% & 65.27\% & 58.48\% & 46.69\% & 40.73\% & \underline{53.75\%} & \textbf{53.62\%} & 33.00\% \\

SCoT & 82.64\% & 74.62\% & 74.34\% & 68.40\% & 70.73\% & 65.65\% & 49.58\% & 45.33\% & 50.31\% & 49.86\% & \underline{45.83\%} \\

ICoT & \textbf{92.26\%} & {82.05\%} & \textbf{81.51\%} & {74.36\%} & \underline{76.58\%} & \textbf{72.95\%} & {55.05\%} & \underline{51.63\%} & {52.11\%} & \underline{52.83\%} & \textbf{50.50\%} \\

\midrule
\multicolumn{12}{l}{\textbf{DeepSeekCoder-6.7B-Instruction}} \\
\midrule

zero-shot & 59.53\% & 50.90\% & 49.62\% & 44.62\% & 48.11\% & 47.70\% & 33.13\% & 32.73\% & 23.28\% & 23.26\% & 29.33\% \\

few-shot & \underline{86.60\%} & {78.65\%} & {74.15\%} & {70.06\%} & \textbf{82.56\%} & \textbf{78.24\%} & \textbf{59.84\%} & \textbf{55.92\%} & \textbf{55.55\%} & \textbf{55.87\%} & \underline{49.67\%} \\

\cmidrule{2-12}

Self-CoT & 81.23\% & 74.62\% & 69.72\% & 66.73\% & 37.04\% & 36.10\% & 24.49\% & 23.75\% & 52.73\% & 52.32\% & 48.83\% \\

ZS-CoT & 73.40\% & 69.36\% & 63.40\% & 61.67\% & 38.31\% & 38.00\% & 25.44\% & 25.01\% & 50.47\% & 50.87\% & 41.50\% \\

SP & 72.64\% & 65.38\% & 61.89\% & 58.46\% & 55.45\% & 52.60\% & 38.64\% & 36.42\% & 36.88\% & 37.25\% & 39.67\% \\

SCoT & {85.94\%} & \underline{80.19\%} & \underline{76.13\%} & \textbf{73.27\%} & {75.44\%} & {71.63\%} & {53.82\%} & {50.17\%} & \underline{54.61\%} & {53.70\%} & {48.33\%} \\

ICoT & \textbf{87.64\%} & \textbf{81.09\%} & \textbf{76.89\%} & \underline{72.63\%} & \underline{78.27\%} & \underline{75.62\%} & \underline{56.51\%} & \underline{54.35\%} & 53.83\% & \underline{55.07\%} & \textbf{51.33\%} \\

\midrule
\multicolumn{12}{l}{\textbf{DeepSeek-V3-671B}} \\
\midrule

zero-shot & 90.94\% & 91.03\% & 80.75\% & 82.82\% & \underline{92.80\%} & \underline{90.80\%} & \textbf{68.73\%} & \underline{65.95\%} & 56.25\% & 57.97\% & 40.67\% \\

few-shot & 96.23\% & 88.46\% & {85.28\%} & 78.46\% & \textbf{93.60\%} & \textbf{91.29\%} & \textbf{68.73\%} & \textbf{66.12\%} & 53.75\% & 54.49\% & {64.00\%} \\

\cmidrule{2-12}

Self-CoT & {97.74\%} & {95.38\%} & {85.66\%} & {86.15\%} & 90.47\% & 87.16\% & 65.96\% & 63.20\% & \underline{59.06\%} & \textbf{60.29\%} & 68.67\% \\

ZS-CoT & \textbf{98.49\%} & \underline{95.64\%} & \textbf{88.68\%} & \textbf{87.95\%} & 90.47\% & 87.55\% & \underline{66.25\%} & 63.69\% & \textbf{59.69\%} & \underline{59.71\%} & {62.67\%} \\

SP & 90.94\% & 88.46\% & 81.13\% & 80.00\% & 87.20\% & 83.53\% & 62.25\% & 58.79\% & 57.19\% & 58.55\% & 63.33\% \\

SCoT & 97.74\% & 95.38\% & 86.79\% & 85.38\% & 82.76\% & 80.55\% & 58.91\% & 56.91\% & 58.12\% & 57.97\% & \underline{74.67\%} \\

ICoT & \underline{98.11\%} & \textbf{97.18\%} & \underline{87.17\%} & \underline{87.44\%} & 83.71\% & {81.71\%} & 60.58\% & 58.40\% & 58.13\% & 59.13\% & \textbf{80.00\%} \\
\bottomrule
\end{tabular}
\caption{Pass@1 results on the simple subset across three models and six code generation benchmarks for difficulty-aware evaluation using Qwen3-8B and GPT-4o. Each model is evaluated under two direct generation strategies (zero-shot and few-shot), four structured prompting baselines (Self-CoT, Zeroshot-CoT, Self-planning, and SCoT), and the proposed method ICoT. Bold and underline indicate the best and second-best among ICoT and all baselines.
}
\label{tab:qwen3-gpt-easy-comparison}
\end{table*}

\paragraph{Results and Analysis.} We further conduct a focused analysis by evaluating all methods exclusively on the subset of problems classified as Simple by Qwen3-8B and GPT-4o. As shown in the main text, Qwen3-8B identifies 53, 275, and 64 simple problems in HumanEval, MBPP-sanitized, and OpenEval, respectively, while GPT-4o identifies 78, 363, and 69 for the same benchmarks. Table~\ref{tab:qwen3-gpt-easy-comparison} presents the pass@1 results on these simple subsets, as determined by the difficulty classifiers using Qwen3-8B and GPT-4o. Despite minor variations in the classification outputs of the two models, the overall performance trends remain consistent.

Across all models and benchmarks, Table~\ref{tab:qwen3-gpt-easy-comparison} reveals an empirical trend: direct generation methods generally outperform most structured prompting baselines on the Simple subset. While ZS-CoT retains relatively strong performance due to its concise reasoning format, other structured methods tend to suffer from noticeable degradation. For instance, on MBPP-sanitized with DeepSeekCoder-6.7B-Instruct under GPT4o routing, Self-CoT achieves only 36.10\%, significantly below the zero-shot baseline of 47.70\% and the few-shot baseline of 78.24\%. Similarly, SP on McEval with Qwen2.5-Coder-3B-Instruct records 33.00\%, underperforming both the zero-shot result of 36.17\% and the few-shot result of 43.33\%. On MBPP-ET with DeepSeek-V3-671B under GPT-4o routing, SCoT scores 56.91\%, which is lower than both the zero-shot score of 65.95\% and the few-shot score of 66.12\%. These results empirically validate a key limitation: indiscriminate use of complex prompting strategies often leads to overthinking on simple tasks, resulting in reduced performance and increased computational cost.

In stark contrast, our proposed ICoT method demonstrates competitive performance across three models. On DeepSeekCoder-6.7B-Instruction, ICoT maintains accuracy levels that are comparable to the few-shot baseline, while on both Qwen2.5-Coder-3B-Instruct and DeepSeek-V3-671B, it achieves either the best or second best results across multiple benchmarks. This suggests that the design of ICoT, which guides the model to abstract task intention rather than prescribing rigid procedural steps, enables it to retain robustness on simple problems. This makes ICoT a reliable component within RoutingGen, effectively minimizing performance penalties even when the router assigns problems that could be reasonably interpreted as either simple or complex.

\subsection{Complex Subset}
\begin{table*}[htbp]
\centering
\small
\setlength{\tabcolsep}{1.5mm}
\begin{tabular}{@{}cccccccccccc@{}}
\toprule
\multirow{2}{*}{\textbf{Method}}  & \multicolumn{2}{c}{\textbf{HumanEval}} & \multicolumn{2}{c}{\textbf{HumanEval-ET}} & \multicolumn{2}{c}{\textbf{MBPP-sanitized}} & \multicolumn{2}{c}{\textbf{MBPP-ET}} & \multicolumn{2}{c}{\textbf{OpenEval}} & \textbf{McEval} \\
& Qwen & GPT & Qwen & GPT & Qwen & GPT & Qwen & GPT & Qwen & GPT & - \\
\midrule
\multicolumn{12}{l}{\textbf{Qwen2.5-Coder-3B-Instruct}} \\
\midrule

zero-shot      
& \textbf{71.53\%} & \textbf{73.08\%} & \underline{63.29\%} & \underline{63.14\%} & 39.38\% & 31.17\% & 25.95\% & 19.61\% & \underline{26.18\%} & \underline{24.59\%} & 13.00\% \\

few-shot       & 64.19\% & 63.95\% & 58.96\% & 57.44\% & \underline{53.22\%} & \underline{46.02\%} & \underline{35.07\%} & \underline{27.42\%} & 23.95\% & 23.72\% & 12.75\% \\

\cmidrule{2-12}

Self-CoT       & 64.23\% & 64.88\% & 58.42\% & 58.08\% & 49.61\% & 41.17\% & 33.12\% & 25.47\% & 23.51\% & 22.75\% & 10.75\% \\

ZS-CoT   & 68.47\% & 68.84\% & 62.52\% & 61.69\% & 50.20\% & 41.41\% & 32.96\% & 25.94\% & 25.83\% & 24.54\% & 15.00\% \\

SP  & 65.23\% & 67.27\% & 57.07\% & 57.50\% & 32.73\% & 26.48\% & 19.28\% & 15.39\% & 25.48\% & 24.27\% & 14.50\% \\

SCoT           & 56.98\% & 56.80\% & 51.08\% & 49.71\% & 45.13\% & 38.75\% & 27.93\% & 22.27\% & 24.17\% & 23.26\% & \underline{21.00\%} \\

ICoT           & \underline{69.86\%} & \underline{72.62\%} & \textbf{64.10\%} & \textbf{65.52\%} & \textbf{55.59\%} & \textbf{47.34\%} & \textbf{36.88\%} & \textbf{31.33\%} & \textbf{26.49\%} & \textbf{24.86\%} & \textbf{22.00\%} \\

\midrule
\multicolumn{12}{l}{\textbf{DeepSeekCoder-6.7B-Instruction}} \\
\midrule

zero-shot      & 39.46\% & 41.45\% & 34.82\% & 35.06\% & 42.40\% & 36.88\% & 27.43\% & 21.87\% & 13.07\% & 12.61\% & 16.00\% \\

few-shot       & \underline{66.04\%} & \textbf{67.27\%} & \textbf{59.55\%} & \textbf{59.01\%} & \underline{56.81\%} & \underline{45.94\%} & \underline{36.81\%} & \underline{27.34\%} & \underline{27.98\%} & \underline{26.51\%} & 24.25\% \\

\cmidrule{2-12}

Self-CoT       & 62.12\% & 62.56\% & 55.77\% & 54.42\% & 36.09\% & 40.08\% & 22.96\% & 25.08\% & 25.57\% & 24.59\% & 26.25\% \\

ZS-CoT   & 59.01\% & 58.49\% & 51.85\% & 50.06\% & 37.47\% & 38.05\% & 23.52\% & 23.28\% & 25.31\% & 23.90\% & 25.75\% \\

SP  & 52.34\% & 53.02\% & 47.97\% & 47.03\% & 37.86\% & 29.84\% & 24.84\% & 18.44\% & 18.60\% & 17.52\% & 18.00\% \\

SCoT           & 63.56\% & 62.27\% & 56.53\% & 53.43\% & 53.72\% & {45.47\%} & 33.49\% & 26.25\% & 26.97\% & 26.28\% & \underline{29.75\%} \\

ICoT           & \textbf{66.22\%} & \underline{65.93\%} & \underline{59.37\%} & \underline{58.14\%} & \textbf{58.52\%} & \textbf{46.41\%} & \textbf{40.62\%} & \textbf{31.02\%} & \textbf{29.47\%} & \textbf{27.57\%} & \textbf{31.25\%} \\

\midrule
\multicolumn{12}{l}{\textbf{DeepSeek-V3-671B}} \\
\midrule

zero-shot      & 83.06\% & 80.70\% & 76.58\% & 73.49\% & \textbf{82.11\%} & \textbf{78.75\%} & \underline{52.76\%} & \underline{46.56\%} & \textbf{41.75\%} & \underline{40.00\%} & 22.00\% \\

few-shot       & 79.28\% & 81.40\% & 71.17\% & 73.26\% & \underline{81.32\%} & \underline{77.50\%} & \textbf{53.29\%} & \textbf{46.88\%} & 36.67\% & 35.41\% & 33.00\% \\

\cmidrule{2-12}

Self-CoT       & \underline{88.29\%} & \textbf{87.67\%} & \underline{79.82\%} & \underline{77.67\%} & 72.24\% & 65.94\% & 49.47\% & 42.50\% & 37.89\% & 36.15\% & \textbf{58.00\%} \\

ZS-CoT   & 87.21\% & 86.51\% & 79.28\% & 72.21\% & 74.87\% & 70.00\% & 49.87\% & 41.87\% & 30.53\% & 29.17\% & 41.00\% \\

SP  & 75.32\% & 73.02\% & 69.37\% & 66.98\% & 68.42\% & 63.44\% & 44.47\% & 39.69\% & 34.74\% & 32.84\% & 34.00\% \\

SCoT           & 87.75\% & 86.98\% & 78.74\% & \underline{77.67\%} & 73.03\% & 72.19\% & 47.11\% & 42.19\% & 40.88\% & \textbf{40.18\%} & 38.00\% \\

ICoT           & \textbf{89.19\%} & \underline{87.44\%} & \textbf{80.54\%} & \textbf{78.37\%} & 75.26\% & 75.00\% & 48.42\% & 44.06\% & \underline{41.23\%} & {39.82\%} & \underline{48.00\%} \\

\bottomrule
\end{tabular}
\caption{Pass@1 results on the complex subset across three models and six code generation benchmarks for difficulty-aware evaluation using Qwen3-8B and GPT-4o. Each model is evaluated under two direct generation strategies (zero-shot and few-shot), four structured prompting baselines (Self-CoT, Zeroshot-CoT, Self-planning, and SCoT), and the proposed method ICoT. Bold and underline indicate the best and second-best among ICoT and all baselines.
}
\label{tab:qwen3-gpt-complex-comparison}
\end{table*}

\paragraph{Results and Analysis.} We further examine model performance on the subset of problems classified as complex by Qwen3-8B and GPT-4o. As shown in the main text, Qwen3-8B identifies 111, 152, and 114 complex problems in HumanEval, MBPP-sanitized, and OpenEval, respectively, while GPT-4o identifies 86, 64, and 109 for the same benchmarks. Despite these differences in classification, the overall performance trends remain stable. As shown in Table~\ref{tab:qwen3-gpt-complex-comparison}, ICoT consistently outperforms all other prompting baselines in the majority of settings, demonstrating strong robustness across three models. Notably, on DeepSeek-V3-671B, ICoT attains a pass@1 of 89.19\% on HumanEval under Qwen3-8B routing and 78.37\% on HumanEval-ET under GPT-4o routing, outperforming both direct generation and other structured strategies. On DeepSeekCoder-6.7B-Instruction, ICoT achieves 58.52\% on MBPP-sanitized under Qwen3-8B routing, surpassing the zero-shot baseline by 16.12 percentage points and achieving the top performance. On Qwen2.5-Coder-3B-Instruct, ICoT scores 31.33\% on MBPP-ET under GPT-4o routing, significantly exceeding the zero-shot result of 19.61\% and all other prompting baselines.  

These consistent results across diverse models and tasks confirm the effectiveness of ICoT in guiding the model to capture task intent. The two components of ICoT, namely the Specification element that defines the input-output constraints and the Idea element that captures the core algorithmic logic and estimates time complexity, guide code generation toward solutions that preserve structural guidance while explicitly modeling the task’s functional requirements.

\newpage
\section{Case Examples of Difficulty-Aware Dynamic Routing}
Simple:
\begin{tcolorbox}[mybox]
\begin{lstlisting}[language=Python]
"question": "\n\ndef encode_shift(s: str):\n    \"\"\"\n    returns encoded string by shifting every character by 5 in the alphabet.\n    \"\"\"\n    return \"\".join([chr(((ord(ch) + 5 - ord(\"a\")) % 26) + ord(\"a\")) for ch in s])\n\n\ndef decode_shift(s: str):\n    \"\"\"\n    takes as input string encoded with encode_shift function. Returns decoded string.\n    \"\"\"\n",
"routing_result": "Simple",
"routing_reason": "because it involves straightforward character shifting with modular arithmetic and no complex logic or edge cases."
\end{lstlisting}
\rule{\linewidth}{0.4pt}
\begin{lstlisting}[language=Python]
"question": "\n\ndef remove_vowels(text):\n    \"\"\"\n    remove_vowels is a function that takes string and returns string without vowels.\n    >>> remove_vowels('')\n    ''\n    >>> remove_vowels(\"abcdef\\nghijklm\")\n    'bcdf\\nghjklm'\n    >>> remove_vowels('abcdef')\n    'bcdf'\n    >>> remove_vowels('aaaaa')\n    ''\n    >>> remove_vowels('aaBAA')\n    'B'\n    >>> remove_vowels('zbcd')\n    'zbcd'\n    \"\"\"\n",
"routing_result": "Simple",
"routing_reason": "because it involves a simple string manipulation task with a straightforward condition check for vowels."
\end{lstlisting}
\rule{\linewidth}{0.4pt}
\begin{lstlisting}[language=Python]
"question": "\n\ndef below_threshold(l: list, t: int):\n    \"\"\"Return True if all numbers in the list l are below threshold t.\n    >>> below_threshold([1, 2, 4, 10], 100)\n    True\n    >>> below_threshold([1, 20, 4, 10], 5)\n    False\n    \"\"\"\n",
"routing_result": "Simple",
"routing_reason": "because it involves a simple loop or built-in function to check all elements against the threshold."
\end{lstlisting}
\rule{\linewidth}{0.4pt}
\begin{lstlisting}[language=Python]
"question": "\n\ndef sum_to_n(n: int):\n    \"\"\"sum_to_n is a function that sums numbers from 1 to n.\n    >>> sum_to_n(30)\n    465\n    >>> sum_to_n(100)\n    5050\n    >>> sum_to_n(5)\n    15\n    >>> sum_to_n(10)\n    55\n    >>> sum_to_n(1)\n    1\n    \"\"\"\n",
"routing_result": "Simple",
"routing_reason": "because it involves a simple arithmetic formula and no complex data structures or edge cases."
\end{lstlisting}
\rule{\linewidth}{0.4pt}
\begin{lstlisting}[language=Python]    
"question": "\n\ndef correct_bracketing(brackets: str):\n    \"\"\" brackets is a string of \"(\" and \")\".\n    return True if every opening bracket has a corresponding closing bracket.\n\n    >>> correct_bracketing(\"(\")\n    False\n    >>> correct_bracketing(\"()\")\n    True\n    >>> correct_bracketing(\"(()())\")\n    True\n    >>> correct_bracketing(\")(()\")\n    False\n    \"\"\"\n",
"routing_result": "Simple",
"routing_reason": "because it requires a simple linear scan with a counter to track balance."
\end{lstlisting}
\rule{\linewidth}{0.4pt}
\begin{lstlisting}[language=Python]
"question": "\n\ndef derivative(xs: list):\n    \"\"\" xs represent coefficients of a polynomial.\n    xs[0] + xs[1] * x + xs[2] * x^2 + ....\n     Return derivative of this polynomial in the same form.\n    >>> derivative([3, 1, 2, 4, 5])\n    [1, 4, 12, 20]\n    >>> derivative([1, 2, 3])\n    [2, 6]\n    \"\"\"\n",
"routing_result": "Simple",
"routing_reason": "because it involves a straightforward application of the power rule for derivatives, with a linear transformation of the input list."
\end{lstlisting}
\end{tcolorbox}

Complex:
\begin{tcolorbox}[mybox]
\begin{lstlisting}[language=Python]
"question": "\n\ndef monotonic(l: list):\n    \"\"\"Return True is list elements are monotonically increasing or decreasing.\n    >>> monotonic([1, 2, 4, 20])\n    True\n    >>> monotonic([1, 20, 4, 10])\n    False\n    >>> monotonic([4, 1, 0, -10])\n    True\n    \"\"\"\n",
"routing_result": "Complex",
"routing_reason": "because it requires checking both increasing and decreasing order, which involves iterating through the list and comparing adjacent elements, and handling edge cases like empty lists or single-element lists."
\end{lstlisting}
\rule{\linewidth}{0.4pt}
\begin{lstlisting}[language=Python]
"question": "\n\ndef common(l1: list, l2: list):\n    \"\"\"Return sorted unique common elements for two lists.\n    >>> common([1, 4, 3, 34, 653, 2, 5], [5, 7, 1, 5, 9, 653, 121])\n    [1, 5, 653]\n    >>> common([5, 3, 2, 8], [3, 2])\n    [2, 3]\n\n    \"\"\"\n",
"routing_result": "Complex",
"routing_reason": "because it requires converting lists to sets for efficient common element detection, handling duplicates, and returning a sorted list."
\end{lstlisting}
\rule{\linewidth}{0.4pt}
\begin{lstlisting}[language=Python]
"question": "\n\ndef largest_prime_factor(n: int):\n    \"\"\"Return the largest prime factor of n. Assume n > 1 and is not a prime.\n    >>> largest_prime_factor(13195)\n    29\n    >>> largest_prime_factor(2048)\n    2\n    \"\"\"\n",
"routing_result": "Complex",
"routing_reason": "because it requires efficient factorization logic and handling of multiple divisors to find the largest prime factor."

\end{lstlisting}
\rule{\linewidth}{0.4pt}
\begin{lstlisting}[language=Python]
"question": "\ndef search(lst):\n    '''\n    You are given a non-empty list of positive integers. Return the greatest integer that is greater than \n    zero, and has a frequency greater than or equal to the value of the integer itself. \n    The frequency of an integer is the number of times it appears in the list.\n    If no such a value exist, return -1.\n    Examples:\n        search([4, 1, 2, 2, 3, 1]) == 2\n        search([1, 2, 2, 3, 3, 3, 4, 4, 4]) == 3\n        search([5, 5, 4, 4, 4]) == -1\n    '''\n",
"routing_result": "Complex",
"routing_reason": "because it requires counting frequencies, analyzing the relationship between frequency and value, and handling edge cases where no valid integer exists."
\end{lstlisting}
\rule{\linewidth}{0.4pt}
\begin{lstlisting}[language=Python]
"question": "\ndef strange_sort_list(lst):\n    '''\n    Given list of integers, return list in strange order.\n    Strange sorting, is when you start with the minimum value,\n    then maximum of the remaining integers, then minimum and so on.\n\n    Examples:\n    strange_sort_list([1, 2, 3, 4]) == [1, 4, 2, 3]\n    strange_sort_list([5, 5, 5, 5]) == [5, 5, 5, 5]\n    strange_sort_list([]) == []\n    '''\n",
"routing_result": "Complex",
"routing_reason": "because it requires alternating between selecting the minimum and maximum values from the remaining elements, which involves careful iteration and state management."
\end{lstlisting}
\rule{\linewidth}{0.4pt}
\begin{lstlisting}[language=Python]
"question": "\ndef triangle_area(a, b, c):\n    '''\n    Given the lengths of the three sides of a triangle. Return the area of\n    the triangle rounded to 2 decimal points if the three sides form a valid triangle. \n    Otherwise return -1\n    Three sides make a valid triangle when the sum of any two sides is greater \n    than the third side.\n    Example:\n    triangle_area(3, 4, 5) == 6.00\n    triangle_area(1, 2, 10) == -1\n    '''\n",
"routing_result": "Complex",
"routing_reason": "because it requires checking the triangle validity condition and applying Heron's formula, which involves multiple steps and careful handling of floating-point precision."
\end{lstlisting}
\end{tcolorbox}

\section{Case Examples of RoutingGen}
Simple:
\begin{tcolorbox}[mybox]
\begin{lstlisting}[language=Python]
"question": "\n\ndef derivative(xs: list):\n    \"\"\" xs represent coefficients of a polynomial.\n    xs[0] + xs[1] * x + xs[2] * x^2 + ....\n     Return derivative of this polynomial in the same form.\n    >>> derivative([3, 1, 2, 4, 5])\n    [1, 4, 12, 20]\n    >>> derivative([1, 2, 3])\n    [2, 6]\n    \"\"\"\n", "entry_point": "derivative", "canonical_solution": "    return [(i * x) for i, x in enumerate(xs)][1:]\n", "test": "\n\nMETADATA = {}\n\n\ndef check(candidate):\n    assert candidate([3, 1, 2, 4, 5]) == [1, 4, 12, 20]\n    assert candidate([1, 2, 3]) == [2, 6]\n    assert candidate([3, 2, 1]) == [2, 2]\n    assert candidate([3, 2, 1, 0, 4]) == [2, 2, 0, 16]\n    assert candidate([1]) == []\n\n"
"code": "def derivative(xs: list):\n    if not xs:\n        return []\n    derivative_coeffs = []\n    for i in range(1, len(xs)):\n        derivative_coeffs.append(xs[i] * i)\n    return derivative_coeffs"
\end{lstlisting}
\rule{\linewidth}{0.4pt}
\begin{lstlisting}[language=Python]
"question": "\ndef add(lst):\n    \"\"\"Given a non-empty list of integers lst. add the even elements that are at odd indices..\n\n\n    Examples:\n        add([4, 2, 6, 7]) ==> 2 \n    \"\"\"\n", "entry_point": "add", "canonical_solution": "    return sum([lst[i] for i in range(1, len(lst), 2) if lst[i]%2 == 0])\n", "test": "def check(candidate):\n\n    # Check some simple cases\n    assert candidate([4, 88]) == 88\n    assert candidate([4, 5, 6, 7, 2, 122]) == 122\n    assert candidate([4, 0, 6, 7]) == 0\n    assert candidate([4, 4, 6, 8]) == 12\n\n    # Check some edge cases that are easy to work out by hand.\n    \n"
"code": "def add(lst):\n    total = 0\n    for index in range(len(lst)):\n        if index % 2 != 0 and lst[index] % 2 == 0:  # Check if index is odd and element is even\n            total += lst[index]\n    return total"
\end{lstlisting}
\rule{\linewidth}{0.4pt}
\begin{lstlisting}[language=Python]
"question": "\n\ndef below_threshold(l: list, t: int):\n    \"\"\"Return True if all numbers in the list l are below threshold t.\n    >>> below_threshold([1, 2, 4, 10], 100)\n    True\n    >>> below_threshold([1, 20, 4, 10], 5)\n    False\n    \"\"\"\n", "entry_point": "below_threshold", "canonical_solution": "    for e in l:\n        if e >= t:\n            return False\n    return True\n", "test": "\n\nMETADATA = {}\n\n\ndef check(candidate):\n    assert candidate([1, 2, 4, 10], 100)\n    assert not candidate([1, 20, 4, 10], 5)\n    assert candidate([1, 20, 4, 10], 21)\n    assert candidate([1, 20, 4, 10], 22)\n    assert candidate([1, 8, 4, 10], 11)\n    assert not candidate([1, 8, 4, 10], 10)\n\n"
"code": "def below_threshold(l: list, t: int):\n    \"\"\"Return True if all numbers in the list l are below threshold t.\n    >>> below_threshold([1, 2, 4, 10], 100)\n    True\n    >>> below_threshold([1, 20, 4, 10], 5)\n    False\n    \"\"\"\n    for num in l:\n        if num >= t:\n            return False\n    return True"
\end{lstlisting}
\end{tcolorbox}

Complex:
\begin{tcolorbox}[mybox]
\begin{lstlisting}[language=Python]
"question": "from typing import List\n\n\ndef sort_numbers(numbers: str) -> str:\n    \"\"\" Input is a space-delimited string of numberals from 'zero' to 'nine'.\n    Valid choices are 'zero', 'one', 'two', 'three', 'four', 'five', 'six', 'seven', 'eight' and 'nine'.\n    Return the string with numbers sorted from smallest to largest\n    >>> sort_numbers('three one five')\n    'one three five'\n    \"\"\"\n", "entry_point": "sort_numbers", "canonical_solution": "    value_map = {\n        'zero': 0,\n        'one': 1,\n        'two': 2,\n        'three': 3,\n        'four': 4,\n        'five': 5,\n        'six': 6,\n        'seven': 7,\n        'eight': 8,\n        'nine': 9\n    }\n    return ' '.join(sorted([x for x in numbers.split(' ') if x], key=lambda x: value_map[x]))\n", "test": "\n\nMETADATA = {\n    'author': 'jt',\n    'dataset': 'test'\n}\n\n\ndef check(candidate):\n    assert candidate('') == ''\n    assert candidate('three') == 'three'\n    assert candidate('three five nine') == 'three five nine'\n    assert candidate('five zero four seven nine eight') == 'zero four five seven eight nine'\n    assert candidate('six five four three two one zero') == 'zero one two three four five six'\n"
"icot": "1: Specification:  \n  - **Input**: A space-delimited string where each word is a number spelled out in lowercase (e.g., \"three one five\").  \n  - **Output**: A space-delimited string with the same numbers sorted in ascending order (e.g., \"one three five\").  \n  - **Constraints**:  \n    - Valid input words are 'zero' to 'nine'.  \n    - The output must maintain the same format (space-separated words).  \n\n2: Idea:  \n  - **Core Logic**:  \n    1. **Map words to numbers**: Create a dictionary to convert each word to its corresponding integer (e.g., {'zero': 0, 'one': 1, ..., 'nine': 9}).  \n    2. **Split and convert**: Split the input string into a list of words, then map each word to its integer value.  \n    3. **Sort**: Sort the integers in ascending order.  \n    4. **Map back to words**: Convert the sorted integers back to their word representations.  \n    5. **Join into a string**: Combine the sorted words into a space-delimited string.  \n  - **Time Complexity**: **O(n log n)** (due to sorting), where n is the number of words in the input.  \n  - **Space Complexity**: **O(n)** (for storing intermediate lists and mappings)."
"code": "from typing import List\n\ndef sort_numbers(numbers: str) -> str:\n    word_to_num = {\n        'zero': 0,\n        'one': 1,\n        'two': 2,\n        'three': 3,\n        'four': 4,\n        'five': 5,\n        'six': 6,\n        'seven': 7,\n        'eight': 8,\n        'nine': 9\n    }\n    \n    num_to_word = {v: k for k, v in word_to_num.items()}\n    \n    words = numbers.split()\n    nums = [word_to_num[word] for word in words]\n    nums_sorted = sorted(nums)\n    words_sorted = [num_to_word[num] for num in nums_sorted]\n    \n    return ' '.join(words_sorted)"
\end{lstlisting}
\rule{\linewidth}{0.4pt}
\begin{lstlisting}[language=Python]
"question": "\ndef search(lst):\n    '''\n    You are given a non-empty list of positive integers. Return the greatest integer that is greater than \n    zero, and has a frequency greater than or equal to the value of the integer itself. \n    The frequency of an integer is the number of times it appears in the list.\n    If no such a value exist, return -1.\n    Examples:\n        search([4, 1, 2, 2, 3, 1]) == 2\n        search([1, 2, 2, 3, 3, 3, 4, 4, 4]) == 3\n        search([5, 5, 4, 4, 4]) == -1\n    '''\n", "entry_point": "search", "canonical_solution": "    frq = [0] * (max(lst) + 1)\n    for i in lst:\n        frq[i] += 1;\n\n    ans = -1\n    for i in range(1, len(frq)):\n        if frq[i] >= i:\n            ans = i\n    \n    return ans\n", "test": "def check(candidate):\n\n    # manually generated tests\n    assert candidate([5, 5, 5, 5, 1]) == 1\n    assert candidate([4, 1, 4, 1, 4, 4]) == 4\n    assert candidate([3, 3]) == -1\n    assert candidate([8, 8, 8, 8, 8, 8, 8, 8]) == 8\n    assert candidate([2, 3, 3, 2, 2]) == 2\n\n    # automatically generated tests\n"
"icot": "1: Specification:\n  - Input: A non-empty list of positive integers.\n  - Output: The greatest integer that meets the condition (frequency >= integer value), or -1 if none exist.\n\n2: Idea:\n  - **Frequency Calculation**: First, compute the frequency of each integer in the list using a dictionary (O(n) time and space).\n  - **Filter Valid Candidates**: Collect integers where frequency >= integer value (O(n) time).\n  - **Return Maximum Valid Candidate**: Return the maximum from the valid candidates (O(m) time, where m is the number of valid candidates). If no valid candidates, return -1.\n  - **Overall Time Complexity**: O(n) (dominated by frequency calculation).\n  - **Space Complexity**: O(n) (for storing frequencies). \n\n### Optimized Approach:\n1. Use collections.Counter for efficient frequency counting.\n2. Iterate through the frequency dictionary to filter valid candidates.\n3. Return the maximum valid candidate or -1 if none exist. \n\nThis ensures an efficient and clear solution."
"code": "from collections import Counter\n\ndef search(lst):\n    freq = Counter(lst)\n    valid_numbers = [num for num in freq if freq[num] >= num]\n    return max(valid_numbers) if valid_numbers else -1"
\end{lstlisting}
\rule{\linewidth}{0.4pt}
\begin{lstlisting}[language=Python]
"question": "\ndef by_length(arr):\n    \"\"\"\n    Given an array of integers, sort the integers that are between 1 and 9 inclusive,\n    reverse the resulting array, and then replace each digit by its corresponding name from\n    \"One\", \"Two\", \"Three\", \"Four\", \"Five\", \"Six\", \"Seven\", \"Eight\", \"Nine\".\n\n    For example:\n      arr = [2, 1, 1, 4, 5, 8, 2, 3]   \n            -> sort arr -> [1, 1, 2, 2, 3, 4, 5, 8] \n            -> reverse arr -> [8, 5, 4, 3, 2, 2, 1, 1]\n      return [\"Eight\", \"Five\", \"Four\", \"Three\", \"Two\", \"Two\", \"One\", \"One\"]\n    \n      If the array is empty, return an empty array:\n      arr = []\n      return []\n    \n      If the array has any strange number ignore it:\n      arr = [1, -1 , 55] \n            -> sort arr -> [-1, 1, 55]\n            -> reverse arr -> [55, 1, -1]\n      return = ['One']\n    \"\"\"\n", "entry_point": "by_length", "canonical_solution": "    dic = {\n        1: \"One\",\n        2: \"Two\",\n        3: \"Three\",\n        4: \"Four\",\n        5: \"Five\",\n        6: \"Six\",\n        7: \"Seven\",\n        8: \"Eight\",\n        9: \"Nine\",\n    }\n    sorted_arr = sorted(arr, reverse=True)\n    new_arr = []\n    for var in sorted_arr:\n        try:\n            new_arr.append(dic[var])\n        except:\n            pass\n    return new_arr\n", "test": "def check(candidate):\n\n    # Check some simple cases\n    assert True, \"This prints if this assert fails 1 (good for debugging!)\"\n    assert candidate([2, 1, 1, 4, 5, 8, 2, 3]) == [\"Eight\", \"Five\", \"Four\", \"Three\", \"Two\", \"Two\", \"One\", \"One\"], \"Error\"\n    assert candidate([]) == [], \"Error\"\n    assert candidate([1, -1 , 55]) == ['One'], \"Error\"\n\n    # Check some edge cases that are easy to work out by hand.\n    assert True, \"This prints if this assert fails 2 (also good for debugging!)\"\n    assert candidate([1, -1, 3, 2]) == [\"Three\", \"Two\", \"One\"]\n    assert candidate([9, 4, 8]) == [\"Nine\", \"Eight\", \"Four\"]\n\n"
"icot": "1: Specification:  \n  - **Input**: An array of integers (may contain values outside 1-9).  \n  - **Output**: A list of strings where:  \n    - Only integers between 1 and 9 (inclusive) are considered.  \n    - These integers are sorted in ascending order, reversed, and replaced with their corresponding English names (\"One\" to \"Nine\").  \n    - Non-1-9 integers are ignored.  \n    - Empty input returns an empty list.  \n\n2: Idea:  \n  - **Filter**: Extract integers in the range [1, 9] from the input array.  \n  - **Sort**: Sort the filtered list in ascending order (O(n log n) time).  \n  - **Reverse**: Reverse the sorted list (O(n) time).  \n  - **Map to Names**: Replace each digit with its English name using a dictionary lookup (O(n) time).  \n  - **Overall Complexity**: O(n log n) due to sorting.  \n\n"
"code": "def by_length(arr):\n    # Filter numbers between 1 and 9 inclusive\n    filtered = [x for x in arr if 1 <= x <= 9]\n    # Sort the filtered list in ascending order\n    filtered.sort()\n    # Reverse the sorted list\n    filtered.reverse()\n    # Map numbers to their corresponding names\n    number_names = {\n        1: \"One\",\n        2: \"Two\",\n        3: \"Three\",\n        4: \"Four\",\n        5: \"Five\",\n        6: \"Six\",\n        7: \"Seven\",\n        8: \"Eight\",\n        9: \"Nine\"\n    }\n    # Replace each number with its name\n    result = [number_names[num] for num in filtered]\n    return result"
\end{lstlisting}
\end{tcolorbox}
\section{Few-shot Generation Prompt}
\begin{tcolorbox}[mybox]
\begin{lstlisting}[language=Python]
def reverse_String(str):
""" Write a python function to reverse the given string."""

def reverse_String(str):
    reversed_str = ""  
    for ch in str[::-1]:  
        reversed_str += ch  
    return reversed_str  
\end{lstlisting}
\rule{\linewidth}{0.4pt}
\begin{lstlisting}[language=Python]
def sum_Of_Squares(lst):
""" Write a python function to calculate the sum of squares of the numbers in the given list. """  

def sum_Of_Squares(lst):
    total = 0
    for num in lst:
        total += num ** 2
    return total
\end{lstlisting}
\rule{\linewidth}{0.4pt}
\begin{lstlisting}[language=Python]
def binary_Search(sorted_lst, target):
""" Write a python function to perform binary search on a sorted list to find the target element."""   

def binary_Search(sorted_lst, target):
    low = 0
    high = len(sorted_lst) - 1
    while low <= high:
        mid = (low + high) // 2
        if sorted_lst[mid] == target:
            return mid
        elif sorted_lst[mid] < target:
            low = mid + 1
        else:
            high = mid - 1
    return -1
\end{lstlisting}
\end{tcolorbox}

\section{ICoT-Guided Generation Prompt}
\subsection{Stage 1: ICoT Generation}
\begin{tcolorbox}[mybox]
\begin{lstlisting}[language=Python]
def reverse_String(str):
    \"\"\" Write a python function to reverse the given string.\"\"\"
    Pass

Please analyze the requirement and provide a structured solving process with the following format:  
    1: Specification  
      - Define the input and output format concisely.   
    2: Idea: Summarize the core logic and optimal algorithm with time complexity.  

1: Specification:
  - Input: A string.
  - Output: The reversed version of the input string.
2: Idea:
  - Use slicing `[::-1]` to efficiently reverse the string in O(n) time complexity.
\end{lstlisting}

\rule{\linewidth}{0.4pt}
\begin{lstlisting}[language=Python]
def sum_Of_Squares(lst):
    \"\"\" Write a python function to calculate the sum of squares of the numbers in the given list. \"\"\"
    Pass

Please analyze the requirement and provide a structured solving process with the following format:  
    1: Specification  
      - Define the input and output format concisely.  
    2: Idea: Summarize the core logic and optimal algorithm with time complexity.  

1: Specification:
  - Input: A list of numbers (integers or floats).
  - Output: The sum of the squares of the numbers in the list.
2: Idea:
  - Iterate over the list, square each element, and sum the results.
  - This approach has O(n) time complexity, where n is the length of the list.
\end{lstlisting}

\rule{\linewidth}{0.4pt}
\begin{lstlisting}[language=Python]
def binary_Search(sorted_lst, target):
  \"\"\" Write a python function to perform binary search on a sorted list to find the target element. \"\"\"
  Pass

Please analyze the requirement and provide a structured solving process with the following format:  
    1: Specification  
      - Define the input and output format concisely.  
    2: Idea: Summarize the core logic and optimal algorithm with time complexity.  

1: Specification:
  - Input: A sorted list of numbers (integers or floats) and a target value.
  - Output: The index of the target in the list, or -1 if not found.
2: Idea:
  - Use binary search to find the target in O(log n) time.
\end{lstlisting}
\end{tcolorbox}

\subsection{Stage 2: Code Generation}
\begin{tcolorbox}[mybox]
\begin{lstlisting}[language=Python]
def reverse_String(str):
    """ Write a python function to reverse the given string."""
  
    """
    1: Specification:
      - Input: A string.
      - Output: The reversed version of the input string.
    2: Idea:
      -	Use slicing `[::-1]` to efficiently reverse the string in O(n) time complexity.
    """
# Please check the above solving process and write a code based on it. Note that the solving process may contain errors.


def reverse_String(str):
    return str[::-1]
\end{lstlisting}

\rule{\linewidth}{0.4pt}
\begin{lstlisting}[language=Python]
def sum_Of_Squares(lst):
    """ Write a python function to calculate the sum of squares of the numbers in the given list. """  

    """
    1: Specification:
      - Input: A list of numbers (integers or floats).
      - Output: The sum of the squares of the numbers in the list.
    2: Idea:
      - Iterate over the list, square each element, and sum the results.
      - This approach has O(n) time complexity, where n is the length of the list.
    """

# Please check the above solving process and write a code based on it. Note that the solving process may contain errors.

def sum_Of_Squares(lst):
    total_sum = 0  
    for x in lst:  
        total_sum += x ** 2  
    return total_sum  

\end{lstlisting}

\rule{\linewidth}{0.4pt}
\begin{lstlisting}[language=Python]
def binary_Search(sorted_lst, target):
    """ Write a python function to perform binary search on a sorted list to find the target element."""  

    """
    1: Specification:
      -	Input: A sorted list of numbers (integers or floats) and a target value.
      -	Output: The index of the target in the list, or -1 if not found.
    2: Idea:
      -	Use binary search to find the target in O(log n) time.
    """

# Please check the above solving process and write a code based on it. Note that the solving process may contain errors.

def binary_Search(sorted_lst, target):
    low, high = 0, len(sorted_lst) - 1
    while low <= high:
        mid = (low + high) // 2  
        if sorted_lst[mid] == target:
            return mid
        elif sorted_lst[mid] < target:
            low = mid + 1
        else:
            high = mid - 1
    return -1
\end{lstlisting}
\end{tcolorbox}

\end{document}